\documentclass[letterpaper]{article} % DO NOT CHANGE THIS
\usepackage[]{aaai24}  % DO NOT CHANGE THIS
\usepackage{times}  % DO NOT CHANGE THIS
\usepackage{helvet}  % DO NOT CHANGE THIS
\usepackage{courier}  % DO NOT CHANGE THIS
\usepackage[hyphens]{url}  % DO NOT CHANGE THIS
\usepackage{graphicx} % DO NOT CHANGE THIS
\usepackage{natbib}  % DO NOT CHANGE THIS AND DO NOT ADD ANY OPTIONS TO IT
\usepackage{caption} % DO NOT CHANGE THIS AND DO NOT ADD ANY OPTIONS TO IT
\usepackage{algorithm}
\usepackage{algorithmic}

\usepackage{newfloat}
\usepackage{listings}
\DeclareCaptionStyle{ruled}{labelfont=normalfont,labelsep=colon,strut=off} % DO NOT CHANGE THIS
\floatstyle{ruled}
\newfloat{listing}{tb}{lst}{}
\floatname{listing}{Listing}
\usepackage{multirow}
\usepackage{subcaption}
\usepackage{graphicx}
\usepackage{booktabs} % for professional tables
\usepackage{amssymb}
\usepackage{mathtools}
\usepackage{amsthm}

\newcommand{\floor}[1]{\lfloor #1 \rfloor}
\newcommand{\ceil}[1]{\lceil #1 \rceil}

\title{FlexHB: a More Efficient and Flexible Framework for \\Hyperparameter Optimization}
\author{
    Yang Zhang\textsuperscript{\rm 1}, 
    Haiyang Wu\textsuperscript{\rm 1}, 
    Yuekui Yang\textsuperscript{\rm 1,2} 
}
\affiliations{
    \textsuperscript{\rm 1} Technology Engineering Group, Tencent, Beijing, China\\
    \textsuperscript{\rm 2} Department of Computer Science, Tsinghua University, Beijing, China\\
    yizhizhang@tencent.com, gavinwu@tencent.com, yuekuiyang@tencent.com
}

\begin{document}

\maketitle

\begin{abstract}
    Given a Hyperparameter Optimization(HPO) problem, how to design an algorithm to find optimal configurations efficiently? 
    Bayesian Optimization(BO) and the multi-fidelity BO methods employ surrogate models to sample configurations based on history evaluations. More recent studies obtain better performance by integrating BO with HyperBand(HB), which accelerates evaluation by early stopping mechanism. However, these methods ignore the advantage of a suitable evaluation scheme over the default HyperBand, and the capability of BO is still constrained by skewed evaluation results.
    In this paper, we propose FlexHB, a new method pushing multi-fidelity BO to the limit as well as re-designing a framework for early stopping with Successive Halving(SH). Comprehensive study on FlexHB shows that (1) our fine-grained fidelity method considerably enhances the efficiency of searching optimal configurations, (2) our FlexBand framework (self-adaptive allocation of SH brackets, and global ranking of configurations in both current and past SH procedures) grants the algorithm with more flexibility and improves the anytime performance.
    Our method achieves superior efficiency and outperforms other methods on various HPO tasks. Empirical results demonstrate that FlexHB can achieve up to $6.9\times$ and $11.1\times$ speedups over the state-of-the-art MFES-HB and BOHB respectively.     
\end{abstract}

\section{Introduction}
\label{sec: intro}
As Machine Learning becomes the game changer in many real world problems, Hyperparameter Optimization (HPO) has attracted increasing research interests \cite{he2021automl,AutoRlparker2022automated} since the performance of ML models heavily depends on particular settings of their hyperparameters 
% In Deep Learning, it is also well known that the choice of crucial hyperparameters can dominate the final evaluation metric
 \cite{yang2020hyperparameter}. Therefore, HPO methods now play a key role in Automatic Machine Learning(AutoML) pipelines \cite{hutter2019automlbook}. HPO can be quite computational-expensive because one may have to try a lot of hyperparameter configurations to find the optimal solution \cite{he2021automl}.
% and each time of configuration evaluation can require considerable resources. 

To enhance the efficiency of HPO algorithms, 
researchers mainly focus on \textit{sampling strategy} and \textit{evaluation scheme} \cite{huang2020BOSS}, i.e., 
how to generate new configurations and how to evaluate them. 
Instead of randomly sampling in hyperparameter space \cite{bergstra2012randomsearch}, 
Bayesian Optimization(BO) \cite{hutter2011SMBO,BOsnoek2012practical,frazier2018tutorialofBO} 
uses probabilistic model to approximate the relation between configurations and final performances, 
and 
% this model-based sampling strategy 
soon becomes a prevalent method. 
To improve the efficiency of utilizing computation resources, HyperBand \cite{li2017hyperband} framework emerges, which employs the early stopping mechanism of Successive Halving(SH) \cite{karnin2013oldSuccessiveHalving} to accelerate evaluations. Further researches try to adopt advantages of both BO and HyperBand. For example, BOHB \cite{falkner2018bohb} uses TPE surrogate model \cite{HPOAlgosbergstra2011algorithms} as sampling strategy and employs HyperBand as evaluation scheme. The application of HyperBand algorithms also shed a light on the multi-fidelity HPO
% people wonder if we can make use of evaluation metrics
% % of ML model 
% with different fidelities(i.e., evaluation results given different number of resources)
% to boost the HPO process
 \cite{klein2017FABOLAS,poloczek2017multifidelity}. Low-fidelity measurements such as validation metrics on lower budgets in HyperBand, are presumably not as convincing as high-fidelity ones \cite{kandasamy2017multifidelity}, yet shall contribute to tuning process. BOHB ignores multi-fidelity information, only using highest-fidelity results for fitting the surrogate. 
To address the issue, MFES-HB \cite{li2021mfes} 
% is proposed 
% a new sampling strategy
% , which 
fits a weighted ensemble of surrogate models based on multi-fidelity measurements.
% i.e., evaluation results on different budget levels for early-stopping in HyperBand. 
% Meanwhile, a weighted ensemble of multiple surrogate models fitted on each different fidelities is employed, aiming to balance the biases and profits from multi-fidelity information. 
% By doing so, it can exploit multi-fidelity information without being deluded by low-quality ones.

% Though achieving prevailing efficiency compared to other methods, 
However, sampling strategies in these algorithms are still inadequate for mining information from past configurations. MFES-HB only considers a few levels of fidelities that used for early stopping check in HyperBand. 
% leaving out the majority of fidelity levels which are not used for early-stopping check in HyperBand.
% The amount of training resources of multi-fidelity measurements can vary a lot since allocated budgets grow exponentially in HyperBand.
% Thus,
The number of higher-fidelity measurements decreases dramatically, while lower-fidelity measurements are densely collected at similar resource levels.
% which could make more difficult to fit a satisfying ensemble of different fidelities.
As a result, the extracted multi-fidelity information is still limited. 
% \textit{Can we enlarge the multi-fidelity information to better guide the configuration sampling, without adding considerable resource costs?}
\textit{Can we design a new sampling strategy that better utilizes information from history trials, without adding considerable resource costs?
}

% Recent HyperBand-based methods, including BOHB, MFES-HB, DEHB 
% \cite{DEHB-ijcai21}, and HB-ABLR \cite{valkov2018HB-ABLR}, 
% have proposed different sampling strategies, 
% while all complying to the default evaluation scheme of HB. 
% Though there are researches aimed to run SH brackets asynchronously 
% \cite{li2018ASHA} or improve the methodology of SH \cite{huang2020BOSS}, 
% the \textit{arrangement} of different initialized SH is still fixed in 
% the big picture.

To evaluate sampled configurations, HyperBand-based methods \cite{DEHB-ijcai21,valkov2018HB-ABLR} use SH as building blocks, 
and they all suffer from the nature of vanilla SH \cite{li2018ASHA}, 
stopping some potential good configurations by mistake. 
Besides, HyperBand framework is congenitally inefficient due to the design of mechanically looping over 
a fixed sequence of SH subroutines, which is not flexible at all when handling with different ML problems. 
Thus, HyperBand(and the SH) could cause wastes of resources and constrain the speed of finding optimal 
configurations, 
and a novel evaluation scheme is needed to overcome these shortcomings.
% to converge to the optimal configuration quickly.
\textit{Can we give more chances to early-stopped configurations?} 
Besides, 
% to obtain measurements of good configurations with higher fidelity?
\textit{Can we take advantage of SH in a more flexible way 
% novel way, instead of mechanically repeating it in a fixed arrangement, 
to speed up hyperparameter tuning?
}

Based on the analysis of problems above, we propose a novel HPO algorithm called \textbf{FlexHB}, which applies an upgraded version of 
Successive Halving (\textbf{GloSH}) with a more flexible framework (\textbf{FlexBand}) 
other than HyperBand, as well as pushing the limits of multi-fidelity HPO (\textbf{Fine-Grained Fidelity}). 
% After discussing related works and some necessary background knowledge, 
% we demonstrate our new method in ``Proposed Method''. 
Our experiments on a series of HPO tasks prove that FlexHB performs better 
than other popular methods. In particular, our method achieves extreme efficiency while being up to 11.1$\times$ times faster than BOHB, 6.9$\times$ times faster than MFES-HB. 
% We also provide an open-source implementation of FlexHB for reference\footnote{\url{https://anonymous.4open.science/r/FlexHB}}.

\section{Related Work}
\label{sec:Related Work}
Great progress of Machine Learning application in academic and industrial areas \cite{goodfellow2016deeplearningbook,silver2017AlphaGo} has driven up the demand for automatic Hyper-parameter Optimization(HPO) \cite{hutter2019automlbook}.
In HPO, model-based and heuristic methods are two most prominent types of algorithms \cite{yang2020hyperparameter}.
As a typical model-based method, Bayesian Optimization(BO) has shown great efficiency for tuning hyperparameters \cite{white2021bananas,liu2021BOonNLP}. BO samples new configurations with the help of probabilistic surrogate models, 
including Gaussian Process(GP) \cite{BOsnoek2012practical}, Tree Parzen Estimator(TPE) \cite{HPOAlgosbergstra2011algorithms}, Random Forests(RF) \cite{hutter2011SMBO} and Bayesian Neural Networks \cite{springenberg2016bayesianNN}. Heuristic methods such as Evolution Strategy \cite{hansen2016cma-es} and 
Population Based Training \cite{jaderberg2017PBT} are also drawing research interests, these methods search better configurations by the evolution of a group of trials \cite{olson2016tpotgenetic}.

Tuning in high-dimension or complex hyperparameter space can be a tough problem for vanilla BO with GP as the surrogate \cite{eggensperger2013BOcomparison}. Different surrogates like RF in SMAC \cite{SMAC3} could be beneficial in such case \cite{wistuba2021fewshotBO}, and some following researches have further proposed effective improvements \cite{eriksson2019TurBO,cowen2022HEBO}. 
% Detailed explanation of BO can be found in Section \ref{sec:Preliminary}.

Besides methods for sampling optimal configurations, 
there are studies focusing on terminating some unwanted configurations to 
enhance tuning efficiency \cite{baker2017EarlyStopping,golovin2017googlevizermedianstop,domhan2015CurveFittingES}.  
% by means of median stopping rule \cite{golovin2017googlevizermedianstop}, 
% learning curve extrapolation \cite{domhan2015CurveFittingES}, etc. 
A feasible routine for early stopping mechanism is proposed in Successive Halving(SH) 
% based on the bandit problem
 \cite{karnin2013oldSuccessiveHalving,pmlr-v51-SHanother}. It is further inherited by HyperBand(HB) \cite{li2017hyperband} and shows competitive efficiency. Following this idea, BOHB \cite{falkner2018bohb} combines advantages of BO and HyperBand, and soon becomes a popular method in HPO applications \cite{Microsoft_NNI_2021,liaw2018RayTune}. 
% Detailed explanation of SH and HB can be found in Section \ref{sec:Preliminary}.

Meanwhile, many multi-fidelity optimization methods attempt to better utilize low-fidelity evaluations with less resources (e.g., early stopped configurations) that may help to approximate the objective function, including FABOLAS \cite{klein2017FABOLAS}, Dragonfly \cite{kandasamy2020dragonfly}, A-BOHB \cite{klein2020multifidelityNAS} and so on \cite{poloczek2017multiinformationopt,pmlr-v89-sen19anoisebousingmultifidelity,hu2019TSE}.
% FABOLAS \cite{klein2017FABOLAS} samples a subset of training data to obtain low-fidelity evaluations to feed the BO surrogate. 
MFES-HB \cite{li2021mfes} achieves the state-of-the-art performance in recent researches, it builds surrogates for each fidelity, 
and uses an updating weight vector to integrate predictions from surrogates. 
% Details of BO, SH and HB can be found in Section \ref{subsec:BO}, \ref{subsec:SH}, \ref{subsec:HyperBand} respectively. 
% A brief explanation of multi-fidelity measurements in MFES-HB is shown in Section \ref{subsec: Fine-Grained Fidelity}.

\section{Preliminary}
% All papers submitted for publication by AAAI Press must be accompanied by a valid signed copyright form. They must also contain the AAAI copyright notice at the bottom of the first page of the paper. There are no exceptions to these requirements. If you fail to provide us with a signed copyright form or disable the copyright notice, we will be unable to publish your paper. There are \textbf{no exceptions} to this policy. You will find a PDF version of the AAAI copyright form in the AAAI AuthorKit. Please see the specific instructions for your conference for submission details.
\label{sec:Preliminary}
% \subsection{Bayesian Optimization}
\noindent
\textbf{Bayesian Optimization}\hspace{0.6mm}
\label{subsec:BO}
Since it can be quite expensive to evaluate the underlying relation $f$ between hyperparameters $X$ and the final evaluation 
metric $y$, Bayesian Optimization(BO) approximates $f(X)$ using a surrogate model $M$ (e.g., Gaussian Process). 
Following the idea of SMBO \cite{hutter2011SMBO}, BO seeks the optimal configuration in hyperparameter space $\Omega$ by iterating over following steps:\\
(1) fit the probabilistic surrogate model $M: p(f|D)$ using current history measurement dataset $D$.\\
(2) use $M$ to select the most promising configuration. In BO, this is usually made possible by selecting configurations that maximize the acquisition function, i.e., $X^\ast = argmax_{X\in\Omega}( \alpha(X;M))$, where the acquisition function $\alpha$  is formulated based on the chosen surrogate model.\\
(3) evaluate the configuration $X^\ast$ to get the corresponding performance metric $y^\ast$ (e.g., validation error).\\
(4) add evaluation result $(X^\ast, y^\ast)$ into $D$.

In step (2), commonly-used acquisition functions include Expected Improvement(EI), Probability of Improvement(PI) and Upper Confidence Bound(UCB) \cite{BOsnoek2012practical,wilson2018maximizingBOAcqFunc}. 
% which trades off exploitation and exploration for selecting the next configuration. 
% Traditional BO methods always execute complete train and validation process in step (3), which is believed to hinder the speed of converging to the optimal configuration.
% \subsection{Successive Halving}
\noindent
\textbf{Successive Halving}\hspace{0.6mm}
\label{subsec:SH}
% Training and evaluating ML models requires specific resources, which can be defined by running time, epochs for iterating over datasets, numbers of data batches and so on.
Successive Halving(SH) \cite{SuccessiveHalvingli2016efficient} starts from evaluating configurations with minor resources, and keeps selecting those with better metrics and allocates exponentially more resources for them. 
% Instead of evaluating every candidate with full resources, 
% SH algorithm controls budgets for each configurations and cease underperforming ones before they could consume the full resource.
% In other words, these configurations are chosen to be ``early-stopped'', and only a small fraction of candidates can exploit higher-cost resources.
% new comment
% In the beginning, SH uniformly allocates a budget to a group of configurations and collects the evaluation results. Then the bottom ones with low performance are thrown away while top ones are kept. 
% new comment
% This ``rank and drop'' operation repeats till some configurations reach the full resource. 
A typical SH algorithm runs as below:\\
(1) allocate $r$ units of resources for $n$ configurations. \\
(2) rank the corresponding evaluation performance, determine the top $1/\eta$ ones ($\eta$ is usually set to 3).\\
(3) promote the top $1/\eta$ configurations to next round by allocating $\eta$ times more resources (e.g., epochs for training), terminate the rest. By doing so, $n = n / \eta, r = r\eta$.\\
(4) iterate over step (1) to (3) until budget $R$ is reached.

A procedure of running SH from lowest to highest budget is called a SH bracket, which is further used in the HyperBand algorithm.
% More detailed pseudo-code of SH can be found in Algorithm \ref{SH Algo} of Appendix\ref{appendix: SH and HyperBand}.
% \subsection{HyperBand}

\noindent
\textbf{HyperBand}\hspace{0.6mm}
\label{subsec:HyperBand}
% Successive Halving stops underperforming configurations in advance 
% allows only the most promising configurations to execute with full resource, 
% so that the evaluation becomes much less time-consuming. However, 
SH requires the number of configurations $n0$ to evaluate at the beginning step as a user-specified input. Given a finite total budget $B$, usually there is no prior whether we should consider a large number $n$ of configurations with small resource $r$, or small number $n$ of configurations with large resource $r$. This dilemma is also called ``$n$ vs $B/n$ trade-off''. 
% A small $n0$ with a large $r0$ may lead to a waste of resource for some worthless configurations, while a large $n0$ with a small $r0$ may sacrifice some good configurations by mistake since they are terminated before fulfilling their potentials.
HyperBand(HB) \cite{li2017hyperband} 
% , an algorithm built on SH, 
overcomes this defect by considering several different values of $n$. HB consists of two types of loops, the outer loop enumerates different $n$ and the corresponding $r$, 
% and calls SH given $n$ and $r$ 
% (see Algorithm \ref{HyperBand Algo} in Appendix\ref{appendix: SH and HyperBand}), 
while the inner loop is exactly the SH process. 
Table \ref{HyperBand procedure} shows a typical HB process. HyperBand is actually doing a grid search over feasible $n0$ values, these values are carefully chosen so that every SH execution requires similar amount of resources. HB procedure will execute repeatedly until a satisfying configuration is found.
% or allowed time budget is exhausted.
% \vspace{-2mm}
\begin{table}[h]
% \vskip 0.1in
\begin{center}
\resizebox{0.8\columnwidth}{!}{%
\begin{tabular}{|l|l|l|l|l|l|}
\hline
  & $s = 4$ & $s=3$  & $s=2$  & $s=1$  & $s=0$  \\
  
$i$ & $n_i\;\;r_i$ & $n_i\;\;r_i$ & $n_i\;\;r_i$ & $n_i\;\;r_i$ & $n_i\;\;r_i$ \\
\hline
$0$ & $\textbf{81\;\;1}$ &  $\textbf{27\;\;3}$    & $\textbf{9\;\;\;\:9}$     & $\textbf{6\;\;\;\:27}$     & $\textbf{5\;\;\;\:81}$     \\
$1$ & $27\;\;3$ &  $9\;\;\;\:9$   & $3\;\;\;\:27$     & $2\;\;\;\:81$     &      \\
$2$ & $9\;\;\;\:9$ & $3\;\;\;\:27$     &$1\;\;\;\:81$      &      &      \\
$3$ & $3\;\;\;\:27$  & $1\;\;\;\:81$     &      &      &      \\
$4$ & $1\;\;\;\:81$  &      &      &      &     \\
\hline
\end{tabular}}
\end{center}
\caption{A HyperBand procedure with $R=81, \eta=3$. 
\label{HyperBand procedure}
% Columns show the inner loop(SH brackets).
}
% \vspace{-5mm}
% \vskip -0.2in
\end{table}

\section{Proposed Method}
\label{sec:Proposed Method}
% In this section, we introduce main components of our method in detail. 
\subsection{Fine-Grained Fidelity method}
\label{subsec: Fine-Grained Fidelity}
\noindent
\textbf{Multi-Fidelity BO}
% Our new sampling strategy follows the idea of multi-fidelity BO. 
Given a series of resource budgets allocated by Successive Halving, measurements from different fidelity levels could be categorized into $K$ groups: $D_1, ..., D_K$, where $K = \floor{log_{\eta}(R)}+1 $. 
% Measurements in the same group are trained and evaluated with same number of resources.
To make full use of multi-fidelity measurements,
% we have to combine these surrogate $M_i$ to yield a better approximation of the objective function $f$. 
MFES-HB fits surrogate models $M_i$ based on data points in each $D_i$, and uses a weight vector $w$ to integrate and balance the information from each fidelity level (i.e., the prediction of each base surrogate $M_i$), where $w_i \in [0,1]$ and $\sum_i{w_i} = 1$. 
% Thus, $w_i$ describes the contribution of each surrogate model $M_i$ built on fidelity $r_i$.
Surrogates with more accurate and robust predictions shall be assigned with a larger $w_i$, and vice versa. 
To calculate the value of $w_i$, we can analyze the ordering relation between the prediction from low fidelity surrogate model and the measurement from the full fidelity level. The number of inconsistent ordered pairs (also called ``ranking loss'') is defined as:
\begin{equation}
\label{equation: MFES ranking loss}
L(M_i)=\sum_{j=1}^{|D_K|}\sum_{k=1}^{|D_K|}\textbf{1}((\mu_i(X_j) < \mu_i(X_k))\oplus(y_j < y_k)), 
\end{equation}
where $\oplus$ is the exclusive-or operator, $|D_K|$ is the number of measurements in $D_K$, $(X_i, y_i)$ is full fidelity measurements in $D_K$, and $\mu_i$ is the mean of predictions from $M_i$. The final weight vector is defined as $w_i = \frac{p_i^\gamma}{\sum_{k=1}^{K}{p_k^\gamma}}$, where $p_i$ is the percentage of consistent ordered pairs calculated as $p_i = 1 - \frac{L(M_i)}{\#\ of\ pairs\ in\ D_K}$, and $\gamma$ is usually set to 3.

% \vspace{2}
% \noindent
% \textbf{Analysis of Multi-Fidelity Measurements}
The effect of the multi-fidelity BO above boils down to the quality of measurements in $D_1, ..., D_K$. As a consequence of Successive Halving, we have $|D_1| > |D_2| > ... > |D_K|$, while the resource $r_i$ for $D_i$ is $\eta$ times of $r_{i-1}$ for $D_{i-1}$. We now discuss the characteristics of measurements that leads to deficiency of the algorithm:\\
(1) During tuning, total number of measurements increases slowly. Available data for fitting surrogates and computing weights is limited, especially in the early stage of tuning.
\\
(2) 
Multi-fidelity measurements are distributed unevenly. Roughly speaking, measurements are plentiful for low fidelities while sparse for the full fidelity. In general, surrogate $M_i$ fitted on measurements $D_i$ with larger resource can provide a better approximation of $f$.
% because the resource allocated for those measurements are closer to full fidelity resource $R$.
Biased distributed measurements will not only produce an incompetent full fidelity surrogate $M_K$, but also make difficulty for obtaining proper $w_i$ to integrate surrogates.

\begin{figure}[h]
% \centering
% \vskip 0.1in
\begin{subfigure}[t]{0.25\columnwidth}
\centering
\includegraphics[width=\linewidth]{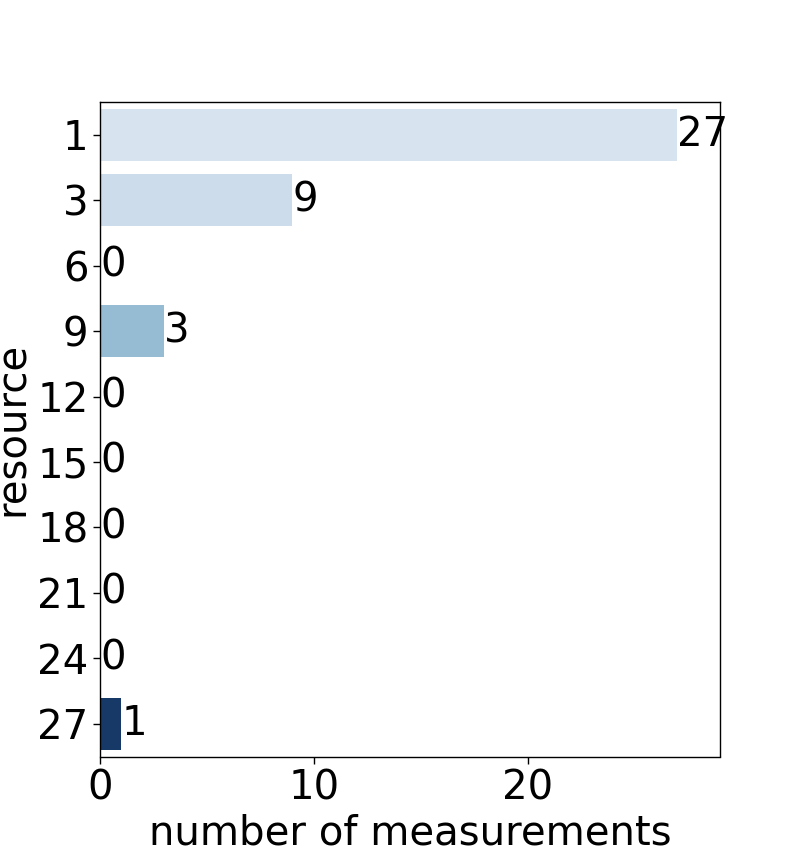}
\caption{vanilla,r0=1}
\label{subfig:vanillan27r1}
\end{subfigure}%
% \hfill
\begin{subfigure}[t]{0.25\columnwidth}
\centering
\includegraphics[width=\linewidth]{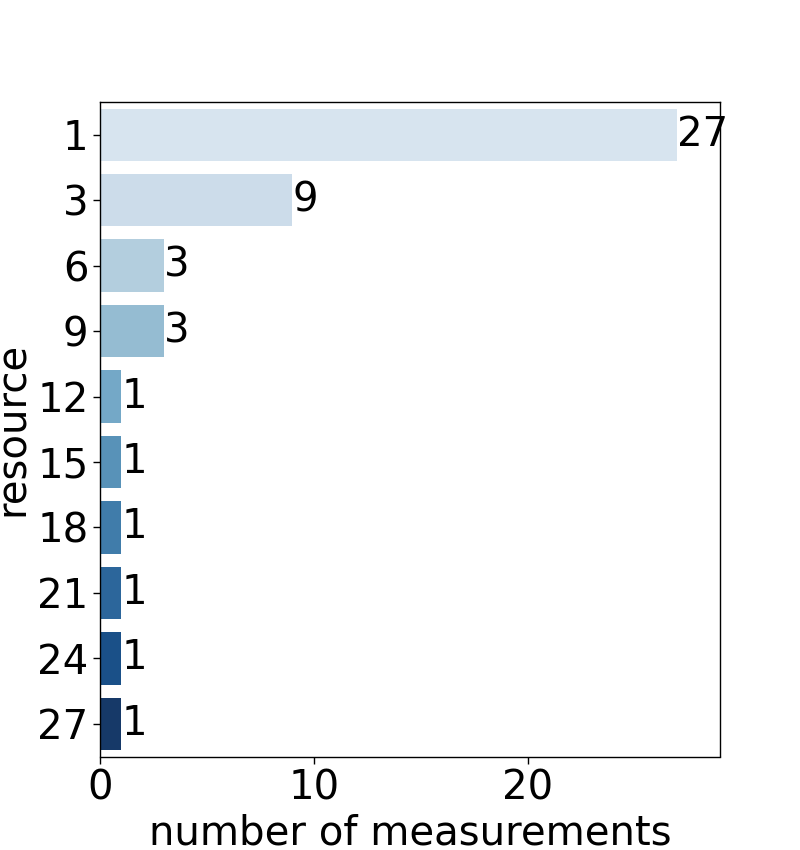}
\caption{FGF,r0=1}
\label{subfig:FGn27r1}
\end{subfigure}%
\begin{subfigure}[t]{0.25\columnwidth}
\centering
\includegraphics[width=\linewidth]{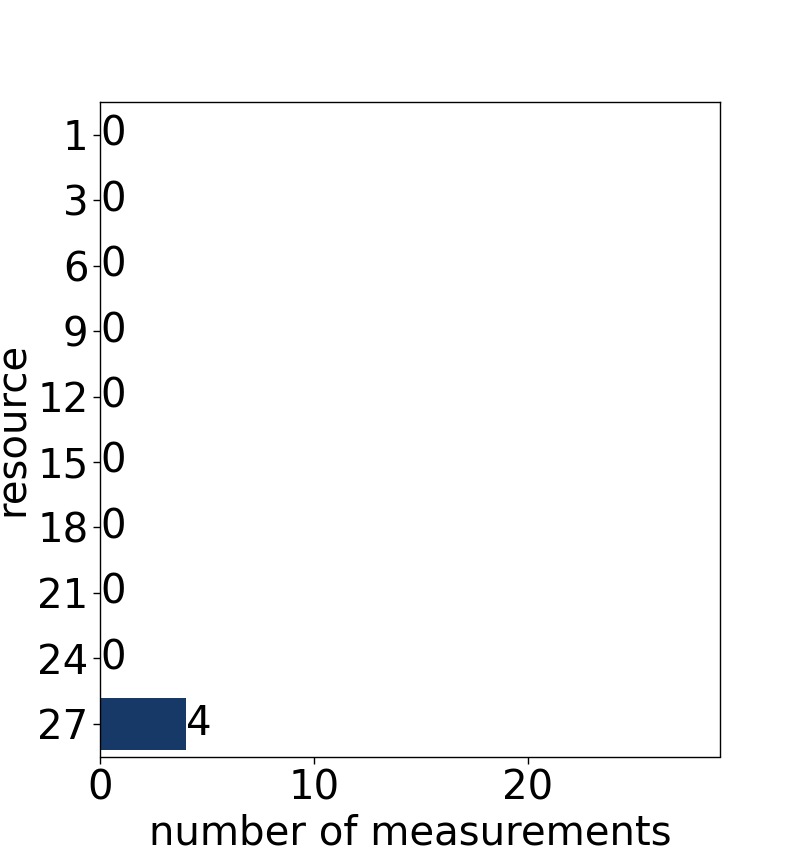}
\caption{vanilla,r0=27}
\label{subfig:vanillan4r27}
\end{subfigure}%
% \hfill
\begin{subfigure}[t]{0.25\columnwidth}
\centering
\includegraphics[width=\linewidth]{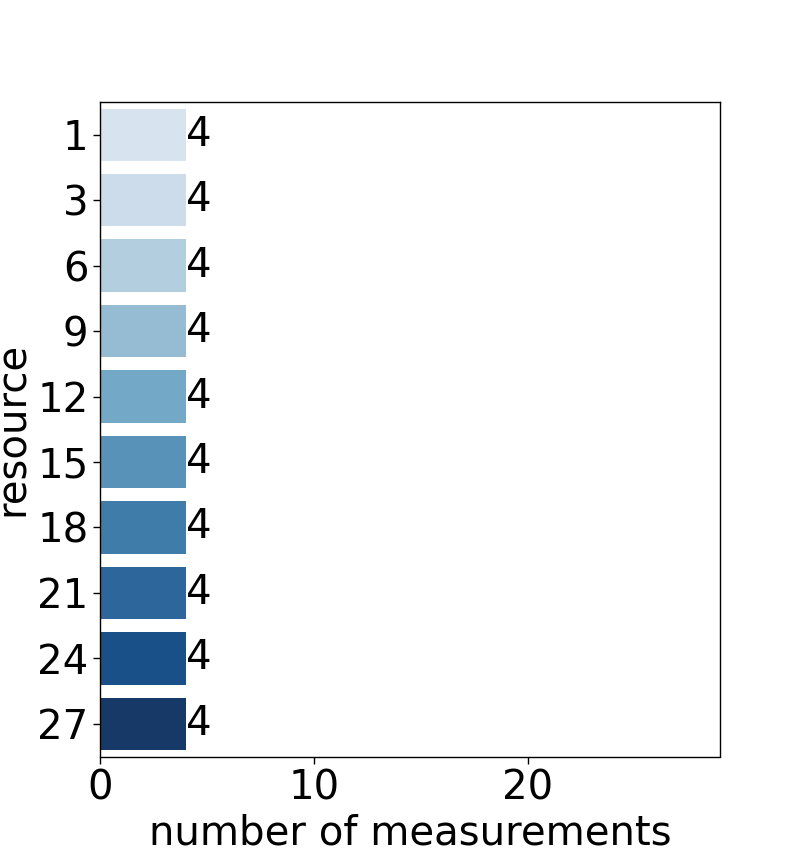}
\caption{FGF,r0=27}
\label{subfig:FGn4r27}
\end{subfigure}
\caption{Measurement distribution over fidelity levels in a bracket. Darker color for higher fidelity, $x$-axis for number of measurements, $y$-axis for resources (different fidelities). 
% Figure \ref{subfig:vanillan27r1},\ref{subfig:vanillan4r27} for vanilla multi-fidelity, and \ref{subfig:FGn27r1},\ref{subfig:FGn4r27} for FGF.
}
\label{fig:Fine Grained Illustration}
% \vskip -0.1in
% \vspace{-2mm}
\end{figure}

Figure \ref{fig:Fine Grained Illustration} shows the distribution of measurements over fidelity levels, Figure\ref{subfig:vanillan27r1} for the most exploring bracket and Figure\ref{subfig:vanillan4r27} for the most exploiting bracket.
% Number of resources is aligned uniformly in the graph (except for the beginning level $r=1$).
We can find that there are no measurements for the majority of resource levels, and most measurements are concentrated in lower fidelity($r=1,3$).
For another example, after a loop of SH brackets (Table \ref{HyperBand procedure}), we have $|D_1|=81, |D_2|=54, |D_3|=27, |D_4|=15, |D_5|=10$.
% Compared to the time cost of completing such a loop, collected measurements is insufficient.
Measurements for the full fidelity ($D_5$) are far less than those for lowest fidelity ($D_1$).

\noindent
\textbf{Fine-Grained Fidelity}
To push the multi-fidelity method to the limit, we propose the \textbf{F}ine-\textbf{G}rained \textbf{F}idelity(\textbf{FGF}) method. The key difference between FGF and traditional multi-fidelity method 
% is that FGF collects additional measurements from various fidelity levels, other than only taking early stopping positions in SH into consideration.
is that FGF collects measurements \textit{uniformly} on various fidelity levels during configuration training.
It collects measurements for every $g$ resource units, 
% (with the additional evaluation at lowest fidelity $r=1$)
and $g$ is set to value of $\eta$ used in SH and HyperBand. 
Therefore, while vanilla multi-fidelity method collects measurements only at early stopping checkpoints in SH executions
(i.e., $r=1,3,9,27,...$ when $\eta=3$), 
FGF evaluates the objective $f$ with \textit{linearly increasing} resources
(i.e., $r=1,3,9,12,15,18,21...$ when $\eta=3$), 
% even if not at the exponentially growing fidelity level used for early stopping selection.
% FGF collects additional measurements that can be used 
% Given a SH bracket with $\eta=3$, $n0=1$ and $r0=27$, measurements utilized by traditional multi-fidelity method 
% are evaluated at $r=1,3,9,27$. 
% Measurements of FGF, as described above, are obtained with $r=1,3,9,12,15,18,21,24,27$. 
and further builds surrogates $M_{r=1}, M_{r=3}, M_{r=9}, M_{r=12}...$ for different $r$.

% In Figure \ref{fig:Fine Grained Illustration}, we can find the difference between vanilla method and FGF. 
Given the same SH bracket, FGF (Figure\ref{subfig:FGn27r1},\ref{subfig:FGn4r27}) obtains more measurements for various fidelity levels 
than vanilla method (Figure \ref{subfig:vanillan27r1},\ref{subfig:vanillan4r27}). 
Note that FGF does NOT increase the total budgets of a bracket, 
instead, it increases the number of times to simply evaluate the current running candidate at some intermediate fidelity levels. 
In most Machine Learning tasks, 
the cost of evaluation is negligible compared to training. 
Thus, FGF enlarges the number of available measurements with tolerable increase of computation cost. 
More importantly, FGF exploits more higher fidelity measurements, 
which are more valuable since they contribute more than low-fidelity ones for building a better ensemble to approximate the full fidelity $f$. 
% Though in FGF measurements for lowest fidelity is still the most abundant,
Besides, accumulated numbers of measurements for lower and higher fidelities are much closer 
than that in vanilla Multi-Fidelity method, 
alleviating the capacity imbalance over different $D_i$. 
% As for the most valuable fidelity levels (the full fidelity and its neighborhood), 
% FGF measurements are several times more than vanilla method. 

In FGF, weights of surrogates are calculated like MFES-HB, 
but the formulation for the weight of full fidelity surrogate ($w_K$ for $M_K$) is re-designed. 
Instead of using the ranking loss obtained by cross validation on $D_K$ itself directly, FGF obtains 
the cross validation ranking loss $L^{\prime}_{K-1}$ and $L^{\prime}_{K}$ on $D_{K-1}$ and $D_{K}$, and calculate the \textit{simulated} percentage of consistent ordered pairs of $M_K$ as $p_K = p_{K-1}\cdot\frac{L^{\prime}_{K}}{L^{\prime}_{K-1}}$, where $p_{K-1}$ is obtained by $L_{K-1}$ following 
Formula \eqref{equation: MFES ranking loss}.

% In most Machine Learning tasks, 
% when considering the whole training stage of a model,
% several more times of evaluation will not add much overhead, since 
% the cost of evaluation is often negligible compared to training. Thus, FGF enlarges the number of available measurements with tolerable increase of computation cost. 
% More importantly, FGF exploits more higher fidelity measurements to emphasize surrogates of higher fidelity, 
% which contributes a lot for building a better ensemble to approximate the full fidelity $f$.
% Further discussion of FGF can be found in Section \ref{subsec: Experiments of FGF}.
% A full comparison of measurement distributions, calculation of weights in FGF and other details can be found in Appendix\ref{appendix: MF vs FGF}, \ref{appendix: weights of FGF}.

% \vspace{-2mm}

\subsection{Successive Halving with Global Ranking}
\label{subsec: GloSH}

\begin{figure*}[h]
% \vskip 0.1in
\centering
  \includegraphics[width=0.95\textwidth]{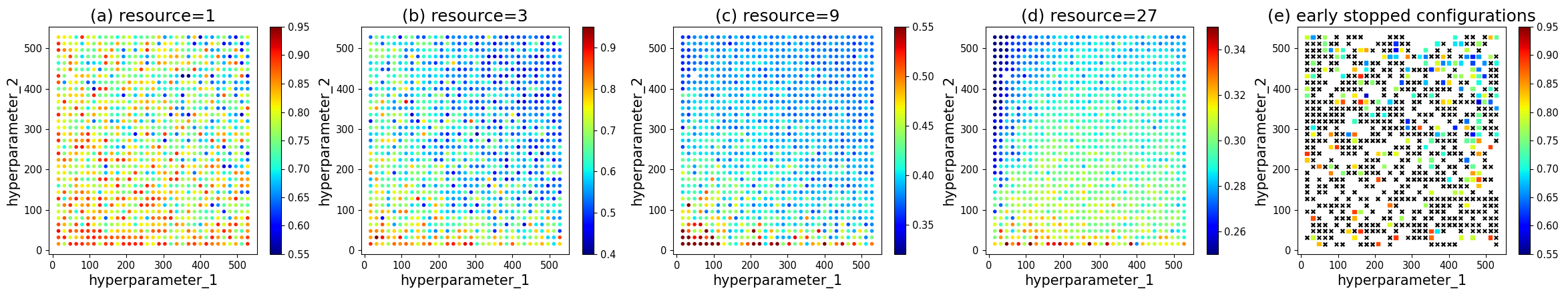}
  \caption{A 2D hyperparameter space for tuning, $x$-axis for values of first hyperparameter and $y$-axis for another. Cooler color for lower validation error points (better configurations), warmer color for higher validation error points. Configurations terminated before reaching the full resource are marked in black ``X'' in Figure \ref{fig:Main Glosh}e. 
  Details 
%   of this hyperparameter space 
  can be found in  Appendix C.3.}
  \label{fig:Main Glosh}
%   \vskip -0.2in
  % \vspace{-4mm}
\end{figure*}
% \noindent
% \textbf{Shortcomings of Successive Halving}
% As discussed in Section \ref{sec:Preliminary}, algorithms based on HyperBand call SH as subroutines for multiple times during tuning.  
% HyperBand method is believed to alleviate this inherent drawback \cite{li2017hyperband} by calling SH with different initializing budgets ($r0$). 
% However, 
% % by investigating early stopping operations of a SH bracket(step (3) in Section \ref{subsec:SH}, or any column in Table \ref{HyperBand procedure}) in detail, we can tell that 
% there are still several times of terminations \textit{after} the beginning budget. Although HyperBand tries to balance the ``$n$ vs $B/n$'' trade-off for the beginning round(bold entries in Table \ref{HyperBand procedure}), it is certainly not able to help with wrongly stopped configurations in subsequent rounds (non-bold in Table \ref{HyperBand procedure}).
% The crux lies in the ``rank and drop'' operation in SH. 
% As stated in Section \ref{subsec:SH}, 
For each early stopping fidelity level, SH ranks current configurations by their metrics and drops 
inferior ones. 
This ``rank and drop'' operation is flawed because:\\
(1) SH is often called for multiple times during tuning, but it only compares observations \textit{locally} in the current SH procedure(i.e., a set of configurations generated in the beginning round), ignoring configurations in past SH procedures. \\
(2) once a configuration is terminated at a fidelity level, it never gets a second chance to continue with more resources. 
However, poor-performing configurations at current fidelity 
might become better ones given higher budget in the future.

Figure \ref{fig:Main Glosh} provides an illustration for this problem. Evaluation results with different fidelity levels are visualized as heat maps.
% where low validation error points (better configurations) are in cool colors,  high validation error points are in warm colors. 
% Figure \ref{fig:Main Glosh}e visualizes the distribution of early-stopped configurations. 
In Figure \ref{fig:Main Glosh}a to \ref{fig:Main Glosh}d , we can find that the upper region of hyperparameter space(especially the upper-left) is the optimal area, and the bottom region(especially the bottom-left) is the worst area. As resources increase, the diversity between configurations become clear gradually. However, Figure \ref{fig:Main Glosh}e shows that SH terminates a considerable proportion of points in good area, while leaving quite a few points in the bottom-left not early-stopped.

% \vspace{2}
% \noindent
% \textbf{Successive Halving with Global Ranking}
To address the shortcoming of vanilla SH, we proposed a redesigned SH algorithm, 
namely \textbf{S}uccessive \textbf{H}alving with \textbf{Glo}bal Ranking (\textbf{GloSH}). 
In GloSH, the ``rank and drop'' operation considers current configurations as well as historic records at 
the corresponding fidelity level, ranks them as a whole to select top ones. 
By doing so, a runing SH procedure can utilize observations from all past procedures. 
% making the early stopping selections more reliable. 
It is worth mentioning that a terminated configuration now is hopefully to be \textit{``revived''} 
in later SH brackets and executed with more resources, as long as it places in the top of the ranking. 

As shown in Algo. \ref{Glosh Algo Main}, we maintain a group of sets $S=\{S_0,...,S_m\}$ containing old early-stopped configurations for each budget level. 
At any $i$-th budget level $r_i$ during any SH running, 
% local configurations evaluated at $r_i$ are combined with $Si$ to form into a \textit{global} set to be ranked. 
local configurations evaluated at $r_i$ are merged with $S_i$ to be ranked \textit{globally}.
GloSH will successively pick configurations from the top ranked candidates until reaching the required number. 
Once an old terminated configuration is selected in this way, it is removed from $S_i$ since it now gets further resources. 
Meanwhile, configurations not picked will be put into $S_i$, waiting for future comparisons.
% Since measurements at low fidelities can be skewed, 
% the global ranking result at lower fidelity level is more likely to bias from the real capability 
% of configurations. 
% Hence 
An additional probability factor $\lambda_i$ is used to control whether or not to pick an old terminated configuration ($\lambda_i$ is not applied for local configurations from current SH procedure). Larger value is assigned to $\lambda_i$ for higher fidelity level. 
% We use $\lambda_0=0.33$, $\lambda_1=0.55$, $\lambda_2=1.0$ while $r_0=1$, $r_1=3$, $r_2=9$ in a typical HyperBand setting with $R=27$. 
With all $\lambda_i=0$, GloSH will degrade to vanilla SH. 
% The detailed GloSH design is shown in Algorithm \ref{Glosh Algo} of Appendix\ref{appendix: GloSH}.
% \vskip -0.05in
\begin{algorithm}[h]
    % \SetKwInOut{Input}{input}\SetKwInOut{Output}{output}
    \begin{algorithmic}      
     \STATE {\bfseries Input:} initial budget $b_0$, maximum budget $b_{max}$, 
    set of $n$ configurations $C = \{c_1, ..., c_n\}$, 
    probabilities for using the old stopped configurations $\{\lambda_0, ..., \lambda_m\}$,
    set of early-stopped configurations for each budget levels $S_0, ..., S_m$
    \STATE {\bfseries Output:} updated $\{S_0, ..., S_m\}$
     \STATE $i = 0$, $b_i = b_0, n_i = |C|$ 
     % $b_i = b_0, n_i = |C|$ \;
    \WHILE{$b_i \leq b_{max}$}
      \STATE Train and evaluate $C$ with $b_i$ resources 
      \STATE $C_{i,global} = Merge(C, S_i)$ 
      % \tcp*[h]{\smaller{merge current and old configurations into one set}} \;
      
      \STATE $C_{i,global} = Rank\_By\_Metric(C_{i,global})$ 
      % \tcp*[h]{\smaller{rank configurations in descending order by their performance}}\;
      
      \STATE Initialize $C_{keep}$, $C_{throw}$ as $\varnothing$ 
      \WHILE{$|C_{keep}| < \floor{n_i / \eta} $}
          \STATE $ c_j$ is the $j$-th element in $C_{i,global}$ 
          \IF{$c_j$ is from $S_i$}
              \IF{$random() < \lambda_i$}
                  \STATE Add $c_j$ to $C_{keep}$ 
              \ENDIF
          \ENDIF
          \IF{$c_j$ is from $C$}
              \STATE Add $c_j$ to $C_{keep}$ 
          \ENDIF
          \STATE $j = j + 1$ 
      \ENDWHILE
      \STATE $C = C_{keep}$, 
      $C_{throw} = C_{i,global} - C_{keep}$ 
      \STATE $S_i = C_{throw}$, 
    %   \STATE 
      $b_i = b_i \eta$, $n_i = \floor{n_i / \eta}$ 
      % $n_i = \floor{n_i / \eta}$ \;
      \STATE $i = i + 1$ 
    \ENDWHILE
  \end{algorithmic}
    \caption{Pseudo-code for GloSH}
    \label{Glosh Algo Main}
  \end{algorithm}
%   \vskip -0.2in

\subsection{FlexBand: a flexible HyperBand framework}
\label{subsec:FlexBand}
% \noindent
% \textbf{HyperBand framework} 
% Several SH brackets with particular $n0$ and $r0$ make up a typical HyperBand procedure. 
% As stated in Section \ref{subsec:HyperBand}, a typical HyperBand procedure is made up of several SH brackets with different $n0$ and $r0$.
% and this procedure shall repeat multiple times with newly sampled configurations until the time budget of tuning is met. 
% Recent HyperBand-based methods, including BOHB, MFES-HB, DEHB 
% \cite{DEHB-ijcai21}, and HB-ABLR \cite{valkov2018HB-ABLR}, 
% have proposed different sampling strategies, 
% while all complying to the default evaluation scheme of HB. 
% Though there are researches aimed to run SH brackets asynchronously 
% \cite{li2018ASHA} or improve the methodology of SH \cite{huang2020BOSS}, 
% the \textit{arrangement} of different initialized SH is still fixed in 
% the big picture.
% In other words, these HyperBand-like algorithms is doing a grid search endlessly over values of $n0$ to balance the $n$ vs $B/n$ trade-off, regardless of actual problem of tuning.
% Instead of rigidly repeating brackets in a fixed arrangement, 
% can we take advantage of SH (and GloSH) in a more flexible way?
Though there are researches aimed to run SH brackets asynchronously 
\cite{li2018ASHA} or improve the methodology of SH \cite{huang2020BOSS}, 
the \textit{arrangement} of different initialized SH is still fixed in 
the big picture. 
% This is quite challenging because the resource needed to reveal the potential of 
% configurations is different for each particular HPO problem. 
% If the quality of configurations is almost revealed only with a small number of resources 
% (e.g., light-weighted neural networks converge quickly after some data batches), 
% then SH brackets with large $r0$ will cause waste of resources. 
% In contrast, if a lot of resources are needed before configurations can show their effects, 
% running SH too aggressively is not preferable (e.g., keep calling SH with smallest $r0$). 
% Extensive experiments on effects of different arrangements are discussed in Section \ref{subsec: Experiments of Arrangements}.
% \vspace{2}
% \noindent
% \textbf{Self-Adaptive Allocation of Brackets}
% The analysis above inspires us to 
To tackle this challenge, we propose the \textbf{FlexBand} approach to allocate brackets \textit{self-adaptively} based 
on real-time results of HPO. In order to determine a proper arrangement, 
we focus on the capability of resource levels for fulfilling the potential of a configuration, 
especially the capability of first early stopping fidelity of SH 
(first row in Table \ref{HyperBand procedure}). 
We resort to the metric ranking of 
configurations. If rankings at two fidelity levels are similar, 
we can expect the capability for fulfilling the potential of a candidate of these two levels are almost 
same. 

In particular, FlexBand utilizes the rank correlation between measurements of same configurations 
given different resources. 
Given a set of measurements $D_i = \{(X_{i,1}, y_{i,1}), ..., (X_{i,n}, y_{i,n})\}$ at $i$-th fidelity 
level $r_i$, 
it finds measurements of these configurations $X_i$ at previous lower fidelity level $r_{i-1}$ in $D_{i-1}$:
% \begin{equation}
% \vspace{-2mm}
% \label{equation:D_i_i-1}
$D_{i;i-1} = \{(X_{i},y_{i-1}) \in D_{i-1} | X_i\ is\ in\ D_i \}$. 
% \vspace{-2mm}
% \end{equation}
Since configurations $X_i$ in $D_i$ and $D_{i;i-1}$ are same, 
it then compares the corresponding $y_i$ and $y_{i-1}$, namely measurements of these $X_i$ at two adjacent 
fidelity levels.
% Given any $D_{i} = \{(X_{i,0},Y_{i,0}), ..., (X_{i,n},Y_{i, n})\}$, $D_{i;i-1} = \{(X_{i,0},Y_{i-1,0}), ..., (X_{i,n},Y_{i-1, n})\}$ collected as above, we can calculate the Kendall-tau correlation coefficient $\tau$ for these two fidelity level $r_1$ and $r_2$. 
% Given any $D_{i}$ and $D_{i;i-1}$ collected as above, 
Then it calculates the Kendall-tau correlation coefficient $\tau$ \cite{abdi2007kendall}
%  for 
% these two fidelity level $r_i$ and $r_{i-1}$: 
as below: 
% For any pair of $(Y_{1, i}, Y_{2, i})$ and $(Y_{1, j}, Y_{2, j})$, if the sorted order of $(Y_{1, i}, Y_{1, j})$ and $(Y_{2, i}, Y_{2, j})$ agrees, then it is said to be a concordant pair, otherwise a disconcordant pair. That is to say, 

For any pair of configurations $X_j$, $X_k$, 
and the corresponding measurement results $y_j$, $y_k$ 
on two different fidelity levels $r_{i}$, $r_{i-1}$. 
If $y_j < y_k$ (or $y_j > y_k$) holds true for both $r_{i}$ and $r_{i-1}$, 
then $X_j$ and $X_k$ make a concordant pair. The coefficient $\tau$ is calculated as:
\begin{equation}
     \tau = \frac{\#\ of\ concordant\ pairs - \#\ of\ disconcordant\ pairs}{\#\ of\ all\ pairs}
\end{equation}
% From the definition, we can tell that $\tau \in [-1, 1]$. 
% Figure  \ref{fig:Kendall Tau} visualizes the ordering of two sets with different values of $\tau$. 

A larger value of $\tau$ indicates greater similarity between rankings of evaluation results on two fidelity levels. 
If the capability of an exploring bracket 
(bigger $n0$ and smaller $r0$) is good enough, or at least close to those of exploiting brackets 
(smaller $n0$ and bigger $r0$), then it is no longer necessary to use the exploiting one. 
% \begin{figure}[h]
%     \centering
%     \includegraphics[width=0.25\textwidth]{Kendall.png}
%     \caption{Different Kendall-tau}
%     \label{fig:Kendall Tau}
% \end{figure}
% With a considerably large $\tau$ for two adjacent fidelity level, 
% FlexBand treat these two levels as interchangeable. 
% During hyperparameter tuning, 
% we can always obtain $\tau$ 
For any two adjacent fidelities $r_i$ and $r_{i-1}$,  
if $\tau$ is large enough (bigger than specified threshold, usually set to 0.55), 
FlexBand \textit{substitute} a more exploiting bracket ($r_0=r_i$) 
with a more exploring SH bracket ($r_0=r_{i-1}$) (see Algo. \ref{Algo:FlexBand Main}). 
Note that FlexBand does NOT modify the total number of brackets of a outer loop,
 it only changes the proportion of different brackets. 

 \begin{algorithm}[ht!]
    \begin{algorithmic}
          % \SetKwInOut{Input}{input}\SetKwInOut{Output}{output}
      \STATE {\bfseries Input:} maximum resource budget $R$, hyperparameter space $\Omega$, history records dataset $D = \{D_{r_1}, ..., D_{r_m}$\} for each fidelity level $r$, the threshold $K_{thres}$, and $\eta$
     \STATE {\bfseries Output:} best configuration
      % with top evaluation performance
        \STATE Initialize $\textbf{N}, \textbf{R}$ as $\varnothing$ \;
        \STATE $s_{max} = \floor{log_{\eta}{R}}, B = (s_{max} + 1)$ \;
        % $m = 0$
        \FOR{$s \in \{s_{max}, s_{max} - 1, ..., 0\}$}
            \STATE $ n = \ceil{\frac{B}{R}\frac{\eta^s}{s+1} }, r=R\eta^{-s} $ \;
            \STATE Append $n$ into $\textbf{N}$, $r$ into $\textbf{R}$ \;
        \ENDFOR
        \STATE Get rank correlation $K$ for each pair of $D_{r_i}, D_{r_{i-1}}$ \;
        \STATE Make a copy of $\textbf{N}$,$\textbf{R}$ as $\textbf{N}^\prime,\textbf{R}^\prime $ \;
        \FOR{$j \in \{ |\textbf{N}|, |\textbf{N}| - 1, ..., 1\}$}
            \STATE $i  = j - 1$ \;
            \STATE $r_i, r_j, n_i, n_j$ are $i$-th and $j$-th element of $\textbf{R}$,$\textbf{N}$  \;
            \IF{$K_{r_i, r_j} > K_{thres}$}
                  \STATE $\textbf{N}^\prime_{j} = n_i$, $\textbf{R}^\prime_{j} = r_i$  \;
                   % $\textbf{R}^\prime_{j} = r_i$
            \ENDIF
        \ENDFOR
        \FOR{$i \in \{0, 1, ..., |\textbf{N}| \}$}
            % $r_i, n_i$ are $i$-th element of $\textbf{R}$,$\textbf{N}$  \;
            \STATE sample $n_i$ configurations from $\Omega$ as set $C$\;
            \STATE run SH or GloSH given $C$, $n_i$ and $r_i$\;
        \ENDFOR
    \end{algorithmic}
      \caption{Pseudo-code for FlexBand}
      \label{Algo:FlexBand Main}
    \end{algorithm}
    
This new evaluation scheme 
% for self-adaptively allocating SH(GloSH) brackets 
has several advantages: 
(1) it inherits benefits from HB, and overcomes the low efficiency of unwarranted resource exploitation, 
(2) it is easy to implement, 
(3) it supports asynchronous execution of SH to facilitate parallel computation.
% For instance,  given history records allocated with 3 and 9 resource units respectively, if the calculated $\tau$ is close to $1$, there is no harm to use the lower fidelity level with less computation costs instead of the higher fidelity level with more computation costs. 
Table \ref{tab: FlexBand cases} shows arrangements of SH generated by FlexBand ($R=81, \eta=3$). 
Note that in the most aggressive case(Table \ref{tab:FlexBand case2}), FlexBand yields an arrangement 
in which the most exploring bracket additionally executes once and 
the most exploiting bracket no longer exists, 
looks like a ``right-shifted'' version of HyperBand. 
% Pseudo-code and further discussions of FlexBand is shown in 
% Algorithm \ref{Algo:FlexBand} of 
% Appendix\ref{appendix: FlexBand}. 
% The $K$ in pseudo-code is exactly $\tau$ described above, and $K_{thres}$ is set to 0.55 in our practice.
% Threshold for $\tau$ is set to 0.55.
% \vspace{-1mm}
\begin{table}[h]
% \vskip -0.075in
    \centering
    \begin{subtable}{0.5\columnwidth}
        % \centering
        % \vskip -0.05in
        \resizebox{\columnwidth}{!}{%
        \begin{tabular}{|l|l|l|l|l|l|}
        \hline
          & $s = 4$ & $s=3$  & $s=2$  & $s=1$  & $s=0$  \\
        $i$ & $n_i\;\;r_i$ & $n_i\;\;r_i$ & $n_i\;\;r_i$ & $n_i\;\;r_i$ & $n_i\;\;r_i$ \\
        \hline
        $0$ & $\textbf{81\;\;1}$ &  $\textbf{27\;\;3}$    & $\textbf{27\;\;3}$     & $\textbf{9\;\;\;\:9}$     & $\textbf{5\;\;\;\:81}$     \\
        
        $1$ & $27\;\;3$ &  $9\;\;\;\:9$   & $9\;\;\;\:9$     & $3\;\;\;\:27$     &      \\
        
        $2$ & $9\;\;\;\:9$ & $3\;\;\;\:27$     &$3\;\;\;\:27$      &   $1\;\;\;\:81$    &      \\
        
        $3$ & $3\;\;\;\:27$  & $1\;\;\;\:81$     &  $1\;\;\;\:81$    &      &      \\
        
        $4$ & $1\;\;\;\:81$  &      &      &      &     \\
        \hline
        \end{tabular}
        }
        % \vspace{-2mm}
        \caption{a possible case}
        \label{tab:FlexBand case1}
    \end{subtable}%
    \begin{subtable}{0.5\columnwidth}
    % \vskip -0.05in
    % \centering
    \resizebox{\columnwidth}{!}{
    \begin{tabular}{|l|l|l|l|l|l|}
    \hline
      & $s = 4$ & $s=3$  & $s=2$  & $s=1$  & $s=0$  \\
      
    $i$ & $n_i\;\;r_i$ & $n_i\;\;r_i$ & $n_i\;\;r_i$ & $n_i\;\;r_i$ & $n_i\;\;r_i$ \\
    \hline
    $0$ & $\textbf{81\;\;1}$ & $\textbf{81\;\;1}$ &  $\textbf{27\;\;3}$    & $\textbf{9\;\;\;\:9}$     & $\textbf{6\;\;\;\:27}$      \\
    
    $1$ & $27\;\;3$ &  $27\;\;3$    & $9\;\;\;\:9$     & $3\;\;\;\:27$     &   $2\;\;\;\:81$   \\
    
    $2$ & $9\;\;\;\:9$ & $9\;\;\;\:9$     &$3\;\;\;\:27$      &   $1\;\;\;\:81$    &      \\
    
    $3$ & $3\;\;\;\:27$  & $3\;\;\;\:27$     &  $1\;\;\;\:81$    &      &      \\
    
    $4$ & $1\;\;\;\:81$  &  $1\;\;\;\:81$     &      &      &     \\
    \hline
    \end{tabular}
    }
    % \vspace{-2mm}
    \caption{the most aggressive case}
    \label{tab:FlexBand case2}
    \end{subtable}
    \caption{Possible arrangements of brackets in FlexBand}
    \label{tab: FlexBand cases}
    % \vskip -0.1in
\end{table}
\subsection{Summary}
% With the FlexBand procedure calling GloSH brackets self-adaptively, FlexHB keeps fitting surrogate models for fine-grained fidelity levels to sample new configurations. 
% In our implementation, a probability $p_r=0.2$ is introduced to control the fraction of randomly sampling configuration as in HyperBand. Thus, in the worst case that FGF surrogates sample wrong configurations, FlexHB is only $1/p_r$ times slower than HB. 
% Pseudo-code of FlexHB is listed in \ref{appendix: Full Algorithm}.
By combining GloSH, FlexBand and FGF into one method, we 
have an efficient and flexible HPO framework, dubbed as \textbf{FlexHB}. 
To the best of our knowledge, it is the first algorithm
% based on SH 
that utilizes information 
\textit{uniformly from the lifecycle of configuration evaluations} in sampling strategy, 
as well as the first multi-fidelity method that employs a HyperBand-like evaluation scheme 
% evolving from SH and HB, 
% which is 
\textit{specifically designed for flexibility}. 
% Most importantly, the combination of GloSH and FlexBand grants our method with efficiency and robustness 
% facing different HPO problems. 
FlexHB is compatible with the theoretical analysis of original SH and HB (illustrated in our Appendix B.6). 
% Also, 
It is easy to implement and can run asynchronously like other parallelized 
methods \cite{klein2020multifidelityNAS}. 
Besides, a probability $p_r=0.2$ is introduced to control the fraction of 
randomly sampling configuration as in HB. Thus, in the worst case that FGF surrogates sample wrong configurations, FlexHB is only $1/p_r$ times slower than HB.

\section{Experiments and Results}
\label{sec:Experiments}
% \vspace{-1mm}
We now investigate the empirical performance of our method on various hyperparameter tuning tasks, 
our aim is to analyze the effects of FlexHB, mainly including: 
(1) FlexHB is more efficient than other popular HPO algorithms in obtaining optimal configurations. 
(2) Fine-Grained Fidelity method outstrips the traditional multi-fidelity method, producing a better ensemble of surrogates. 
(3) GloSH can overcome the disadvantage of SH. 
(4) FlexBand outperforms HyperBand using flexible SH arrangements.
% \vspace{-1mm}
% (4) contributions of each components in FlexHB.
\subsection{Experiment Settings}
\label{subsec:Experiment Settings}
\noindent
\textbf{Algorithm Baselines}\hspace{0.6mm} We compare our proposed method with following HPO algorithms: (1) RS \cite{bergstra2012randomsearch}: the simple random search used as a standard baseline. (2) SMAC \cite{SMAC3}: a popular implementation of BO. (3) Batch-BO \cite{gonzalez2016batchbo}: a parallelized BO method that concurrently evaluating candidates. (4) HB: the original HyperBand, randomly sampling candidate configurations. (5) BOHB \cite{falkner2018bohb}: a combination of HyperBand and vanilla BO. Note that the surrogate in BOHB is fitted on full-fidelity measurements. (6) MFES \cite{li2021mfes}: a combination of HyperBand and multi-fidelity BO. (7) TSE \cite{hu2019TSE}: a multi-fidelity BO method samples smaller subsets of training data and utilizes their low fidelity measurements. (8) BOMS: BO with Median Stop \cite{golovin2017googlevizermedianstop}, a candidate is stopped if its best metric at some step is worse than a median value calculated based on history configurations. (9) BOCF: BO using Curve Fitting method to predict full-fidelity metric for early stopping determination \cite{domhan2015CurveFittingES}.

\noindent
\textbf{Tuning Tasks}\hspace{0.6mm}
HPO tasks used in our experiments mainly includes: 
(1) MLP: Tuning a 2-layer MLP on MNIST. 
(2) ResNet: Tuning ResNet on CIFAR-10. 
% We follow the classic settings, i.e., 40000 images as train set, 10000 images as test set. Random cropping and horizontal flips are applied as data augmentation.
(3) LSTM: Tuning a 2-layer LSTM on Penn Treebank. 
(4) Traditional ML model: Tuning XGBoost on several classification datasets: Covertype, Adult, Pokerhand and 
Bank Marketing \cite{ucidatasets}. 
(5) LCBench \cite{LCBench}: a public HPO benchmark, containing learning curve data for 35 OpenML datasets. 
For MLP, ResNet and XGBoost tasks, 
the classification error on the test set is recorded as 
the validation metric. 
For the LSTM task, the test perplexity is used as the metric. 
We will terminate tuning once the given total time limit is reached 
(more details in Table \ref{tab:HPO tasks setting} of 
Appendix\ref{appendix: Hyperparameter Space}). 
% In all HyperBand-like algorithms, we set $\eta = 3$ as in  \cite{li2017hyperband}. 
In order to obtain convincing results, 
all experiment are conducted for 10 times with various random seeds. 
For each HPO experiment, we track the wall clock time for tuning(including 
time costs for configuration evaluation as well as sampling new hyperparameters), 
and record the metric of the best configuration found so far. 
To fully demonstrate the performance of each algorithm, the mean of metrics 
and the standard error of the mean will be reported.
% \usepackage{multirow}
% \usepackage{graphicx}
% \vspace{-3mm}
% \vspace{-1mm}
\subsection{Experiment Results}
% \vspace{-1.5mm}
\begin{figure*}[h]
% \vskip 0.1in
     \centering
     \begin{subfigure}[b]{0.5\columnwidth}
         \centering
         \includegraphics[width=\textwidth,height=0.65\columnwidth]{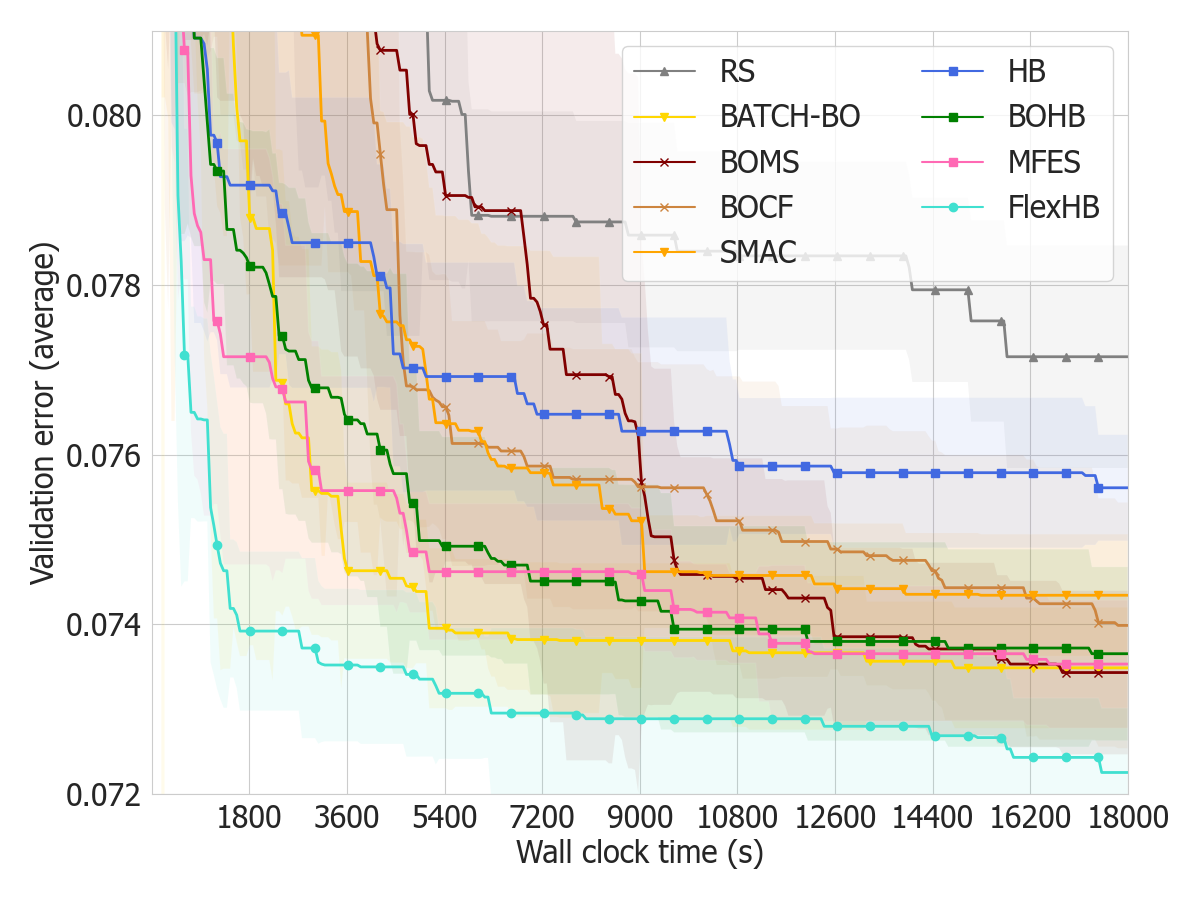}
         \caption{MLP on MNIST}
         \label{fig:MainExp_MNIST}
     \end{subfigure}
     \hfill   
     \begin{subfigure}[b]{0.5\columnwidth}
         \centering
         \includegraphics[width=\textwidth,height=0.65\columnwidth]{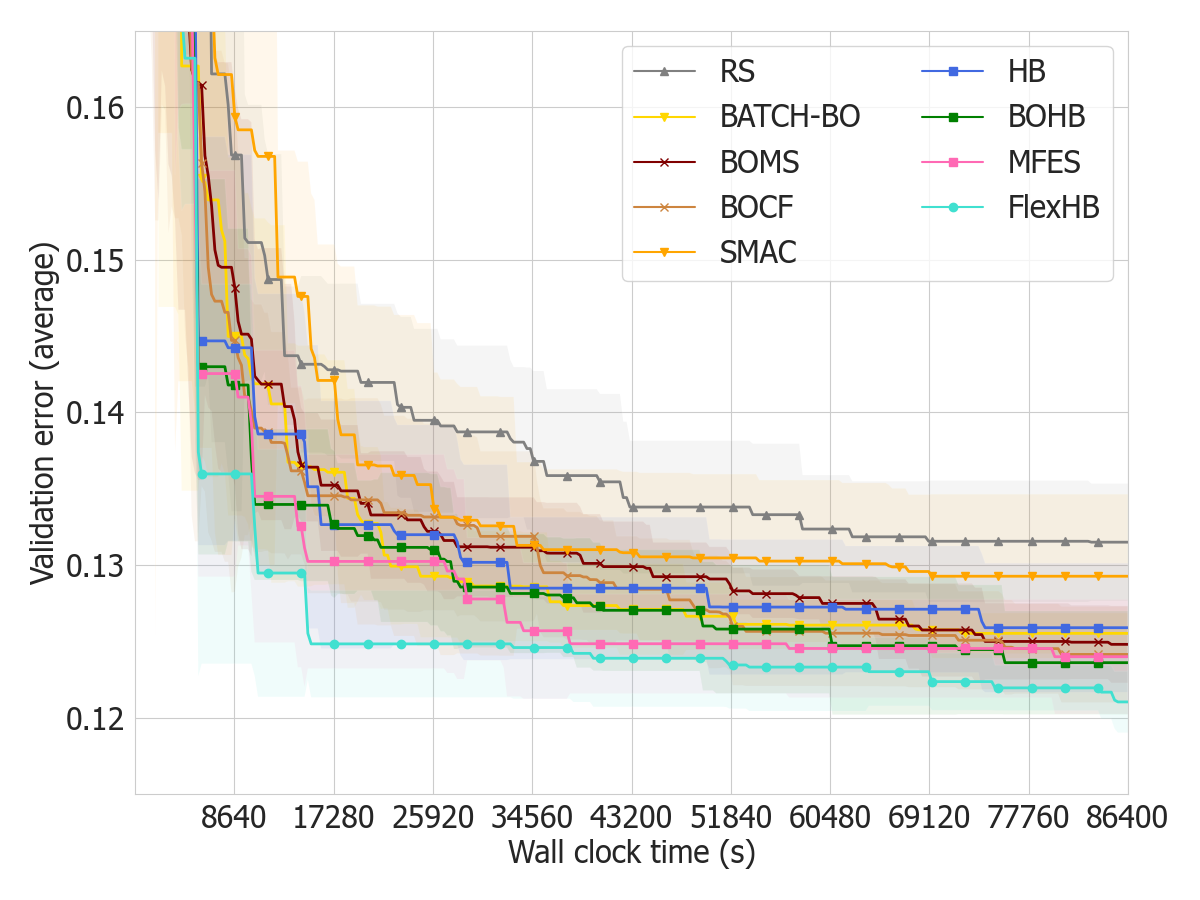}
         \caption{ResNet on CIFAR10}
         \label{fig:MainExp_CIFAR}
     \end{subfigure}
     \hfill   
     \begin{subfigure}[b]{0.5\columnwidth}
         \centering
         \includegraphics[width=\textwidth,height=0.65\columnwidth]{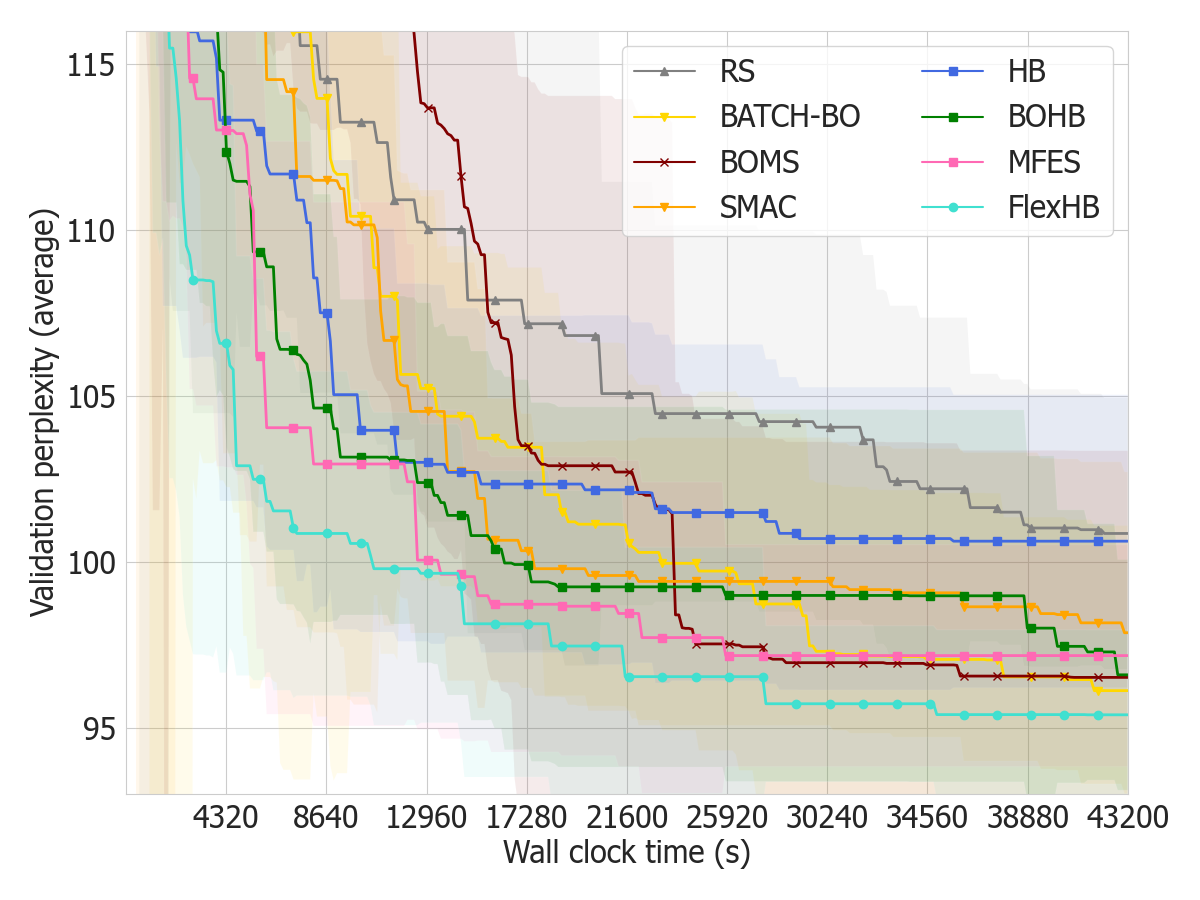}
         \caption{LSTM on Penn Treebank}
         \label{fig:MainExp_PTB}
     \end{subfigure}
     \begin{subfigure}[b]{0.5\columnwidth}
         \centering
         \includegraphics[width=\textwidth,height=0.65\columnwidth]{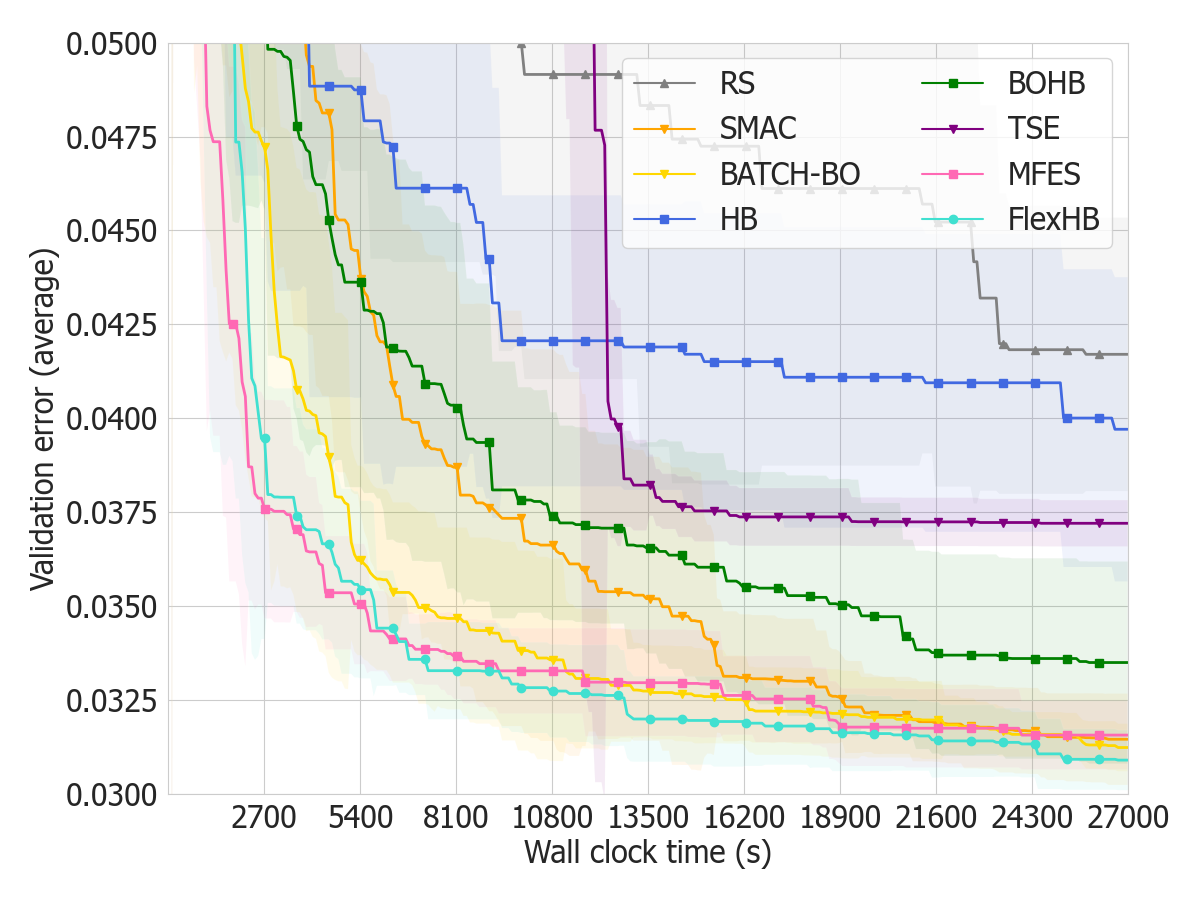}
         \caption{XGBoost on Covtype}
         \label{fig:MainExp_Covtype}
     \end{subfigure}
     \hfill   
     \caption{Results for tuning on four main HPO tasks. LSTM task uses the validation perplexity as metrics.}
     \label{fig:noise level comparison}
    %  \vskip -0.1in
     % \vspace{-4mm}
\end{figure*}

% We now analyze the empirical results of comparing HPO methods on different benchmarks.
% \vspace{-1mm}
\begin{table}[h]
% \vskip 0.1in
\centering
\resizebox{\columnwidth}{!}{%
\begin{tabular}{lcccccc}
\toprule
\multirow{2}{*}{Methods} & \multicolumn{2}{l}{\hfil MLP}                                               & \multicolumn{2}{l}{\hfil ResNet} & \multicolumn{2}{l}{\hfil LSTM} \\ \cline{2-7} 
                         & \multicolumn{1}{l}{Error(\%)}  & \multicolumn{1}{l}{Speed Up}      & Error(\%)   & Speed Up  & Perplexity  & Speed Up \\ \midrule
RS       & 7.72 & F & 13.15 & F  &100.9  & F \\
HB       & 7.56 & 1.0 & 12.59 & 1.0  &100.6  & 1.0 \\
SMAC     & 7.43 & 2.1 & 12.93 & F &97.9& 2.1 \\
BOHB     & 7.36 & 3.7 & 12.36 & 1.5 &96.6  &2.3  \\
BOMS     & 7.34 & 1.9 & 12.48 & 1.1 &96.5& 1.5\\
BOCF     & 7.40 & 1.9 & 12.42 & 1.4 & -  & -\\
Batch-BO & 7.35 & 5.9 & 12.55 & 1.1 &96.1 &1.7\\
MFES     & 7.35 & 5.6 & 12.40 & 2.2 &97.2&2.9\\
FlexHB   & \textbf{7.23} & \textbf{16.1} & \textbf{12.10}& \textbf{4.9} &\textbf{95.4} &\textbf{3.7}\\     
\bottomrule
\end{tabular}%
}
\caption{Converged validation metrics and speed up results (over HyperBand) on different tasks, ``F'' indicates that the method failed to reach the converged metric of HB. Note that BOCF is not applicable to the perlexity of LSTM task.}
\label{tab:speedup}
% \vskip -0.1in
\end{table}
\begin{table}[h]
% \vskip 0.1in
\centering
\resizebox{0.75\columnwidth}{!}{%
\begin{tabular}{lccccc}
\toprule
Methods    & Covtype & Pokerhand & Bank & Adult \\ \midrule
% RS         &         &       &           &      &        \\
SMAC       &  3.15 &1.64   & 8.03  & 12.40   \\
% BatchBO    &  3.12 &           & 8.01  & 12.33   \\
HB         &  3.97 & 3.07 & 8.09  & 12.48   \\
BOHB       &  3.35 & 2.31 & 8.03  & 12.36  \\
% TSE        &         &       &           &      &        \\
MFES       &  3.16 & 1.52 & 8.04  & 12.35  \\
FlexHB &  \textbf{3.09} & \textbf{1.31} &  \textbf{8.00} & \textbf{12.32}       \\
\bottomrule
\end{tabular}%
}
\caption{
    Validation error (\%) on XGBoost tasks. 
    % mean values on 10 times of experiments are listed. 
% Some trivial under-performing methods are omitted. 
Full results are shown in Appendix C.2.}
\label{tab:xgboost result}
% \vskip -0.1in
\end{table}
\noindent
\textbf{HPO Performance}\hspace{0.6mm}
% \noindent
% \textbf{HPO on Neural Networks}
Results of MLP task are shown in Figure \ref{fig:MainExp_MNIST}. 
% RS and HB randomly sample new configurations and become least efficient methods. 
% Early stopping methods including BOMS and BOCF converge slowly in the early stage (within 2 hours) 
% but obtain competitive results in the end. 
% As a powerful multi-fidelity method, MFES outperforms BOHB while still a bit slower than Batch-BO. 
Among all methods, FlexHB achieves best anytime performance as well as best final metric. More precisely, to achieve competitive validation error ($\leq 7.4\%$), it takes 0.45 hours for FlexHB, 1.4 hours for Batch-BO and 3.1 hours for MFES. FlexHB gains a $3.1\times$ speed-up over Batch-BO and $6.9\times$ speed-up over MFES. 
% Table [X] summarizes the test result of methods, indicating that our method can find the optimal configuration compared to others.
Figure \ref{fig:MainExp_CIFAR} shows the results of tuning ResNet on CIFAR-10. 
There are methods like MFES that achieve competent final validation error, yet they failed to converge to the 
(near-)optimal area as quickly as ours. 

In Figure \ref{fig:MainExp_PTB}, we compare these methods by tuning LSTM. Our method achieves prevailing anytime performance, better than any other methods. Table \ref{tab:speedup} summarizes the final metric obtained by these methods on Neural Network tasks and the speed-up over HyperBand (by comparing the time used to exceed the converged metric of HyperBand). Our method shows considerable improvements over other compared algorithms on all tasks. 
More experiments and results on LCBench are detailed in Appendix C.6. 

% \noindent
% \textbf{HPO on XGBoost}
Figure \ref{fig:MainExp_Covtype} shows the result of tuning XGBoost on Covertype dataset. 
% With the worst converged performance, RS and HB are not able to tune quickly. 
MFES achieves greater efficiency especially in the early stage of tuning 
(first 4 hours) compared to SMAC, TSE and BOHB. 
% , showing the advantage of utilizing 
% multi-fidelity measurements. 
After 1.8 hours, FlexHB outperforms all other methods, and finally 
reaches the best converged performance. Final validation errors of tuning XGBoost 
on various datasets are demonstrated in Table \ref{tab:xgboost result}.
% \vspace{-2mm}

\begin{figure}[H]
    % \vspace{-3mm}
     \centering
     \begin{subfigure}[b]{0.495\linewidth}
         \centering
         \includegraphics[width=\textwidth]{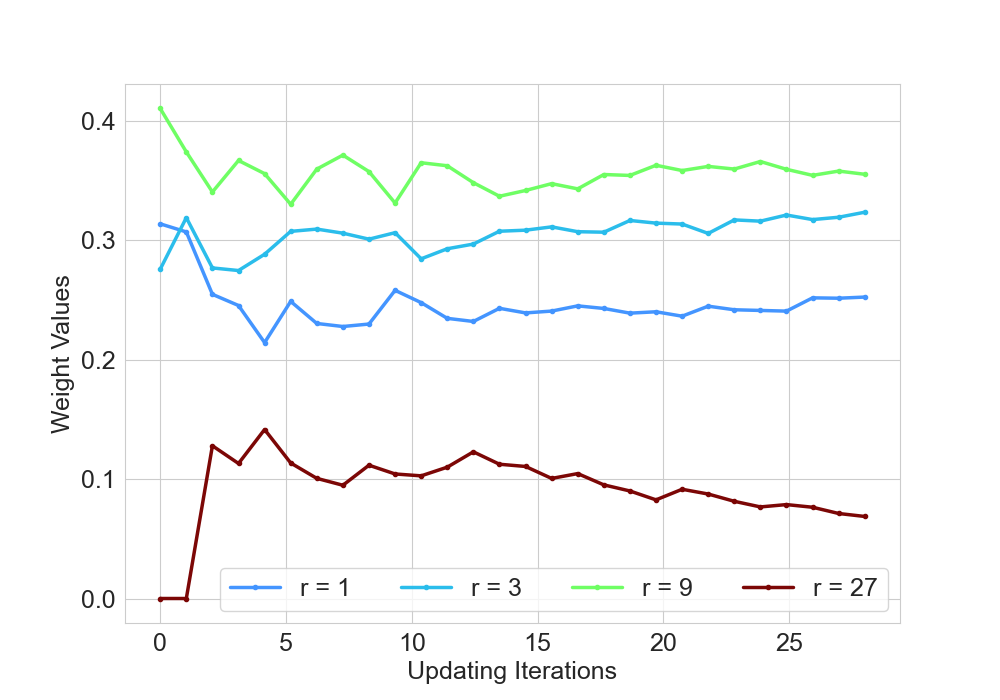}
         \caption{Weights of MFES}
         \label{fig:weights_mfse}
     \end{subfigure}
     \hfill
     \begin{subfigure}[b]{0.495\linewidth}
         \centering
         \includegraphics[width=\textwidth]{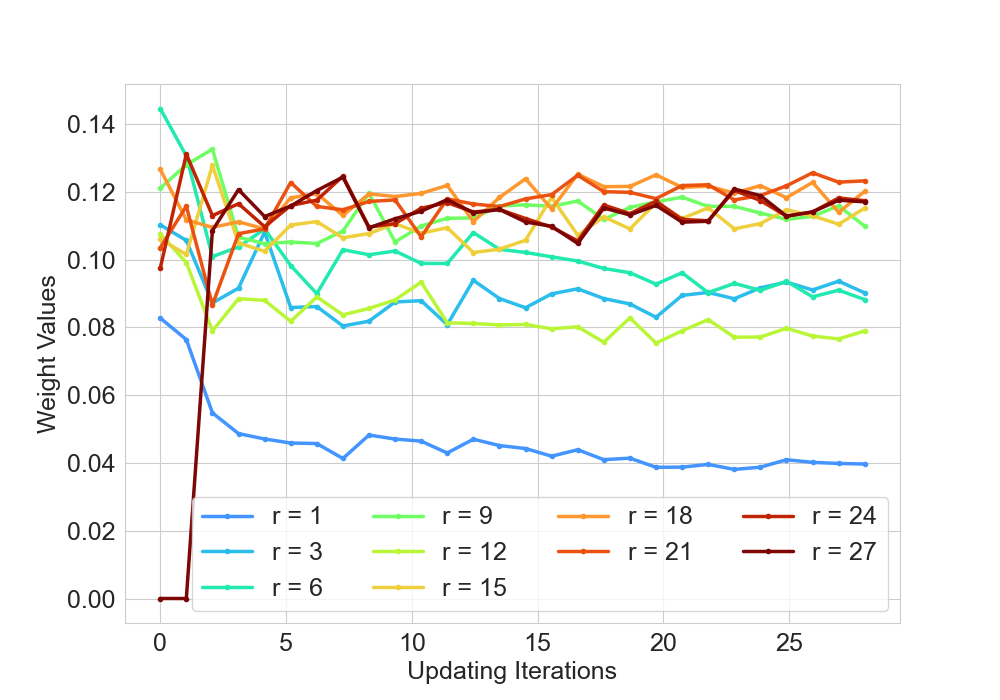}
         \caption{Weights of FlexHB}
         \label{fig:weights_new}
     \end{subfigure}
     \caption{Updating weights in MFES and FlexHB, $x$-axis for iterations and $y$-axis for weight values. 
     Experiments are conducted on the MLP task for 10 times to obtain the stable mean value.}
     \label{fig:weights comparison}
     % \vspace{-0mm}
    %  \vskip -0.15in
\end{figure}

% \subsection{Analysis of FGF weights}
% \subsection{Detailed Analysis}
% \vspace{-2mm}
% In FlexHB, weights of surrogates are calculated like MFES, but we improves the formulation for the weight of full fidelity surrogate ($w_K$ for $M_K$). Instead of using the ranking loss obtained by cross validation on $D_K$ directly, we obtain the cross validation ranking loss $L_{K-1}$ and $L_{K}$ on $D_{K-1}$ and $D_{K}$, and calculate the simulated percentage of consistent ordered pairs of $M_K$ as $p_K = p_{K-1}\cdot\frac{L_{K}}{L_{K-1}}$, where $p_{K-1}$ is obtained by $L_{K-1}$ following Section \ref{subsec: Fine-Grained Fidelity}.
\noindent
\textbf{Analysis of FGF weights}\hspace{0.6mm}
\label{subsec: Experiments of FGF}
% \vspace{-2mm}
Figure \ref{fig:weights comparison} displays all values of $w_i$ in MFES and FlexHB. 
In MFES, weights for lowest and highest fidelity levels($r=1,27$) 
are much smaller than those for middle fidelity levels.
% As tuning progresses, more full fidelity measurements are available and the weight for $r=27$ shall contribute more to final prediction.
The 
weight value for $r=27$ is not increasing significantly while more full fidelity measurements are available 
and the surrogate model for $r=27$ shall contribute more. 
In FlexHB, the weight for lowest fidelity($r=1$) decreases as tuning progresses. 
At the same time, weight for higher fidelity($r=18,21,24,27$) increases and finally becomes 
one of the biggest weights. 
With more higher fidelity measurements collected, the difference between higher and middle fidelity 
gradually grows. 
In summary, the updating of weights in FGF is more reasonable than in vanilla multi-fidelity method. 

\begin{figure}[h]
    % \vspace{-3mm}
    \centering
    \begin{subfigure}[b]{0.4\columnwidth}
         \centering
         \includegraphics[width=\textwidth]{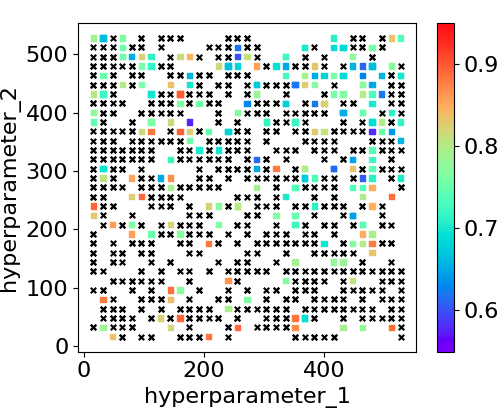}
         \caption{vanilla SH}
         \label{fig:vanillaSHvsGloSH_v4_vanilla}
     \end{subfigure}
     % \hfill
     \begin{subfigure}[b]{0.4\columnwidth}
         \centering
         \includegraphics[width=\textwidth]{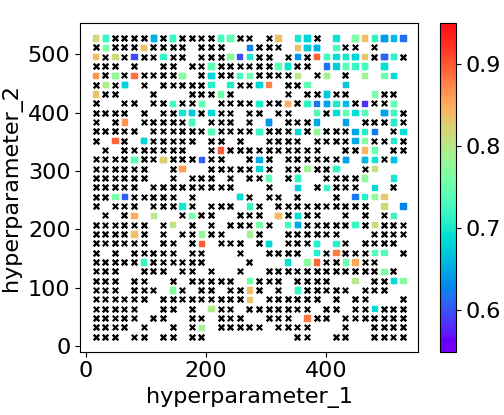}
         \caption{GloSH}
         \label{fig:vanillaSHvsGloSH_v4_glosh}
     \end{subfigure}
    \centering
    \caption{Early stopped configurations in 2D hyperparameter space. 
    %  using SH and GloSH. 
    Early stopped ones are marked in black ``X''.}
    \label{fig:vanillaSHvsGloSH}
    % \vspace{-4mm}
    % \vskip -0.15in
\end{figure}

\noindent
\textbf{Analysis of Glosh}\hspace{0.6mm}
\label{subsec: Experiments of GloSH}
% \vspace{-2mm}
To investigate the effect of GloSH, we visualize all early stopped configurations during 
tuning in Figure \ref{fig:vanillaSHvsGloSH}. 
% As stated in Section \ref{subsec: GloSH}, 
Traditional SH leads to resource wastes in the worst performing area (the bottom left). In contrast, configurations in the worst region are all terminated by GloSH. 
With the global ranking, some early stopped points are ``revived'' in later GloSH brackets, and GloSH obtains more 
full-fidelity measurements for configurations in the better region(the top area), which also helps to fit a better full-fidelity surrogate.

\begin{figure}[h]
    % \vspace{-3mm}
     \centering
     % \begin{subfigure}[b]{0.49\linewidth}
     %     \centering
     %     \includegraphics[width=\textwidth,height=0.6\textwidth]{noiselevel05cmp.png}
     %     \caption{noise level $\phi$ = 0.5}
     %     \label{fig:noise_level05}
     % \end{subfigure}
     % \hfill
     % \begin{subfigure}[b]{0.49\linewidth}
     %     \centering
     %     \includegraphics[width=\textwidth,height=0.6\textwidth]{noiselevel02cmp.png}
     %     \caption{noise level $\phi$ = 0.2}
     %     \label{fig:noise_level02}
     % \end{subfigure}
     \begin{subfigure}[b]{0.49\linewidth}
         \centering
         \includegraphics[width=\textwidth]{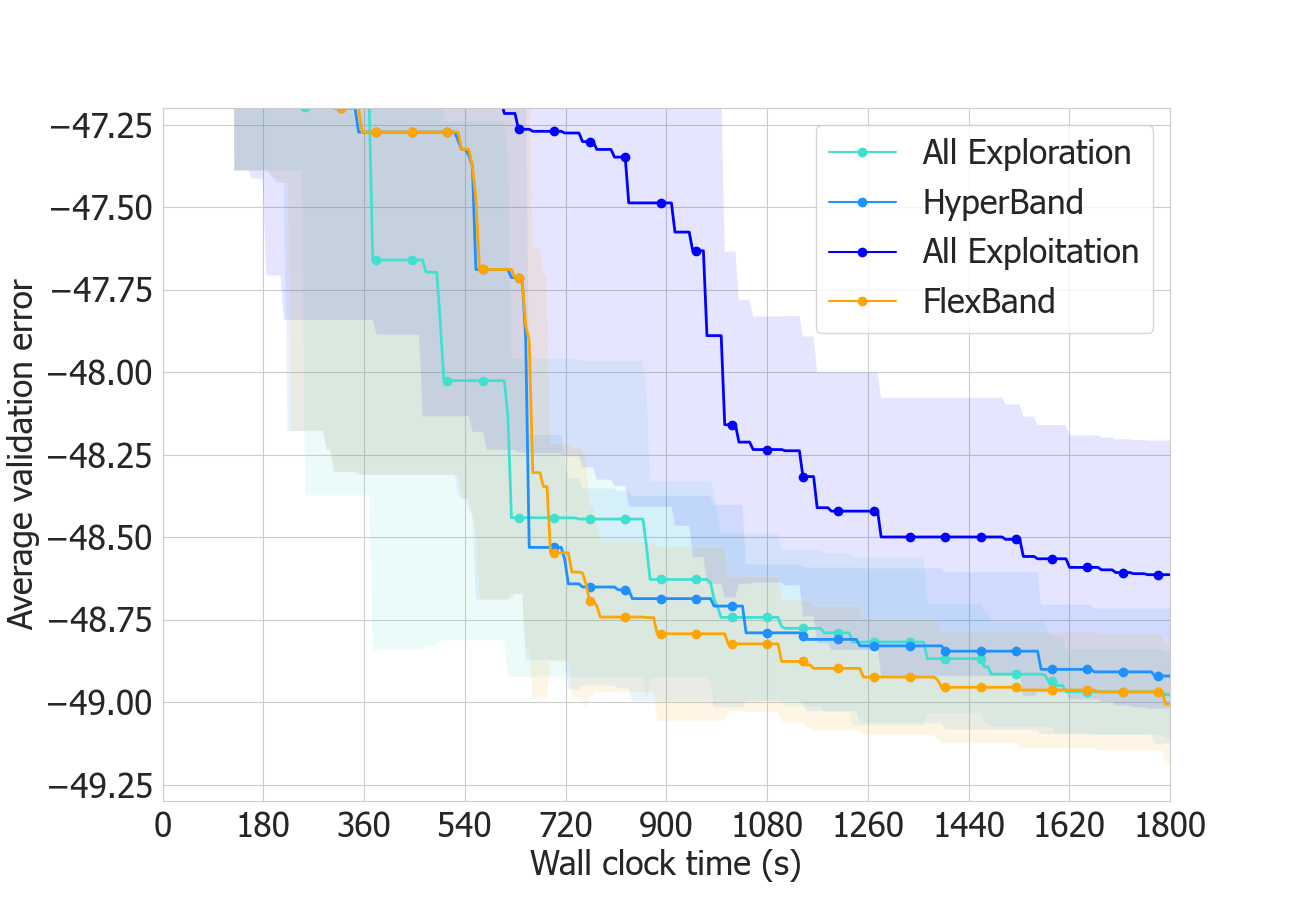}
         \caption{noise level $\phi$ = 0.5}
         \label{fig:curve_noise_level05}
     \end{subfigure}
     \hfill
     \begin{subfigure}[b]{0.49\linewidth}
         \centering
         \includegraphics[width=\textwidth]{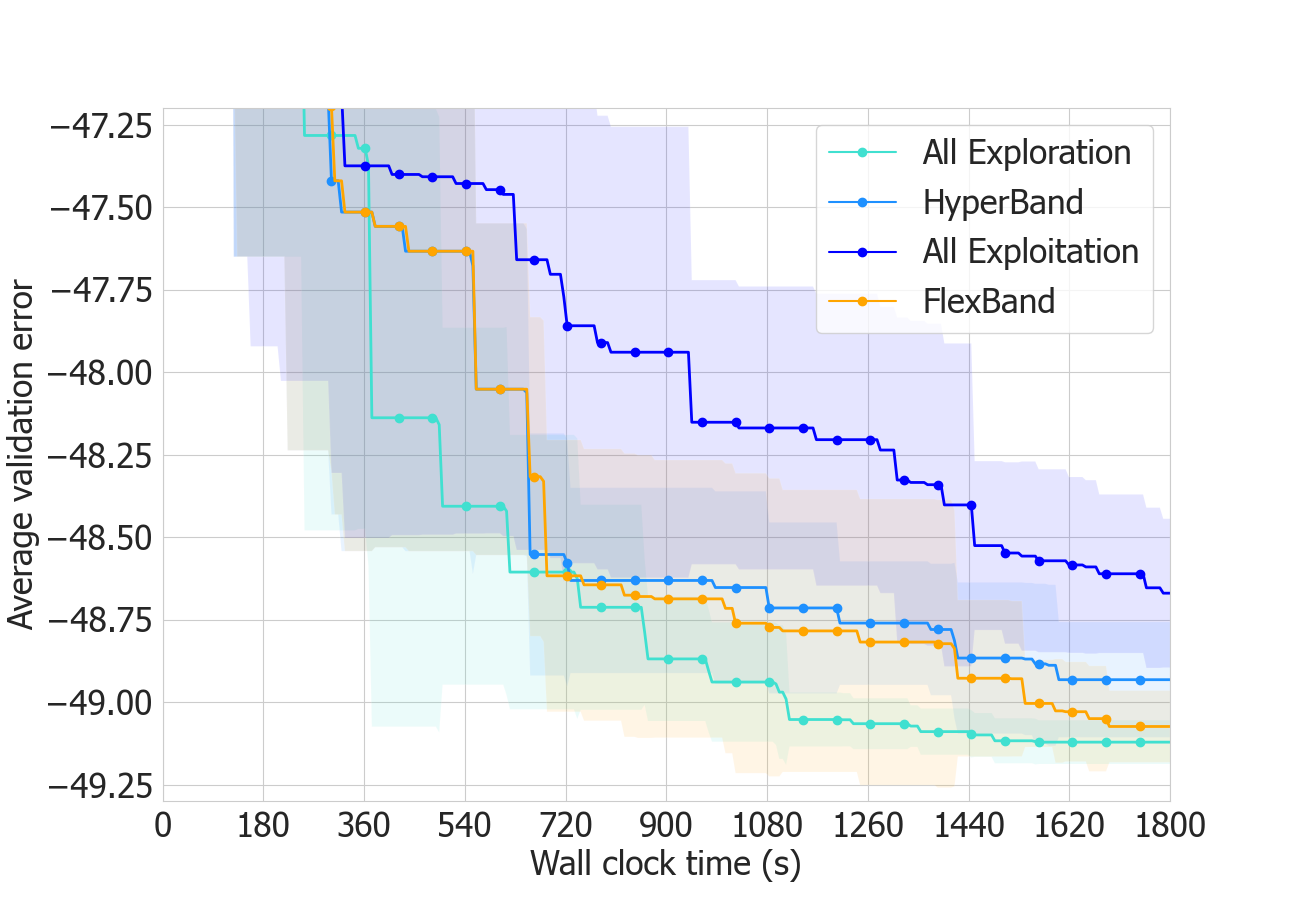}
         \caption{noise level $\phi$ = 0.2}
         \label{fig:curve_noise_level02}
     \end{subfigure}
     \caption{Performance of different SH arrangements under two noise levels.}
     \label{fig:noise level comparison}
     % \vspace{-4mm}
    %  \vskip-0.05in
\end{figure}

\noindent
\textbf{Analysis of SH arrangements}\hspace{0.6mm}
\label{subsec: Experiments of Arrangements}
We fabricate a synthetic data in which a noise factor $\phi$ is used to 
control the variance of validation metrics given a particular configuration 
(see Appendix C.4 for details). With a larger $\phi$, test metric could be more unstable 
% especially in the early stage
especially given small resources, and it is harder to differentiate configurations at low fidelity levels. 
% Figure \ref{fig:noise_level02} and \ref{fig:noise_level05} show learning curves (test error vs the number of resources) of randomly sampled configurations, we can find that 
With a smaller $\phi$ (less variance of metrics), 
relative order of configurations are more consistent among fidelity levels.
For the case with larger variance(Figure \ref{fig:curve_noise_level05}), 
efficiency of HyperBand and All Exploration (all SH brackets initialized with smallest $r0$) are similar. 
But with less variance level (Figure \ref{fig:curve_noise_level02}), 
All Exploration converges to the optimal area with much less time than HyperBand. 
Therefore, although HyperBand always performs better than All Exploitation 
(all SH brackets initialized with largest $r0$), 
it is not yet an efficient arrangement dealing with a variety of situations. 
FlexBand
% , as shown in Figure \ref{fig:noise level comparison}, 
achieves better efficiency than HyperBand in both two cases.

% \vspace{-3mm}
% \subsection{Analysis of GloSH}

% \vspace{-1mm}
% \begin{figure}[h]
%      \centering
%      \begin{subfigure}[b]{0.49\linewidth}
%          \centering
%          \includegraphics[width=\textwidth,height=0.7\textwidth]{curve_noise_level05.png}
%          \caption{noise level $\phi$ = 0.5}
%          \label{fig:curve_noise_level05}
%      \end{subfigure}
%      \hfill
%      \begin{subfigure}[b]{0.49\linewidth}
%          \centering
%          \includegraphics[width=\textwidth,height=0.7\textwidth]{curve_noise_level02.png}
%          \caption{noise level $\phi$ = 0.2}
%          \label{fig:curve_noise_level02}
%      \end{subfigure}
%      % \hfill
%      % \begin{subfigure}[b]{0.49\linewidth}
%      %     \centering
%      %     \includegraphics[width=\textwidth]{curve_noise_level05.png}
%      %     \caption{curve noise level = 0.5}
%      %     \label{fig:curve_noise_level05}
%      % \end{subfigure}
%      \caption{tuning curves with different SH arrangements}
%      \label{fig:noise tuning curves}
% \end{figure}
% We demonstrate the difference between SH arrangements and analyze the benefit of using a proper arrangement. 

\begin{table}[h]
    % \vskip -0.1in
    \centering
    \resizebox{\columnwidth}{!}{%
    \begin{tabular}{lcccccc}
    \toprule
    \multirow{2}{*}{Methods} & \multicolumn{2}{c}{ MLP} & \multicolumn{2}{c}{ ResNet} & \multicolumn{2}{c}{ LSTM} \\ \cline{2-7} & Improve   & $\Delta$(\%)   & Improve    & $\Delta$(\%)    & Improve & $\Delta$(\%) \\ 
    \midrule
    no FGF      & 0.14 & \textbf{-58.8\%} &0.25  &-49.0\%  &3.0  &-42.3\%  \\
    no GloSH    & 0.27 & -20.6\% &0.29  &-40.8\%  &2.0  &\textbf{-61.5\%}  \\
    no FlexBand & 0.24 & -29.4\% &0.18  &\textbf{-63.3\%}  &2.1  &-59.6\%  \\
    FlexHB      & 0.34 & - &0.49  & -  &5.2  &- \\
    \bottomrule
    \end{tabular}
    }
    \caption{Ablation study results. 
    ``Improve'' indicates changes of the converged metric over HyperBand, 
    ``$\Delta$'' indicates the proportion of lost improvement (compared to FlexHB). 
    % In ``improvement'' columns, 
    % Classification error (\%) is used for MLP and ResNet.
    }
    \label{tab:ablation}
    % \vskip -0.05in
\end{table}
% \subsection{Ablation Study}
\noindent
\textbf{Ablation Study}\hspace{0.6mm}
\label{subsec: abl}
% \vspace{-2mm}
We also conduct an ablation study by leaving out one of components in FlexHB. 
For example, the ``no GloSH'' method uses vanilla SH instead of GloSH. 
We compare the final converged metric in Table \ref{tab:ablation}, 
components with the largest gain are shown in bold text. 
In conclusion, each component
% (FGF as sampling strategy, GloSH and FlexBand as evaluation scheme)
leads to performance benefits, and contributions of them varies in different problem settings. 
GloSH gains minor improvement on the MLP task (20.6\%), and we attribute this to the fact that local 
and global ranking of candidates here is more consistent than other tasks. 
In other words, old terminated candidates are less often to rank high and ``revive'', 
constraining the benefits of GloSH. 
% Visualized performances and 
More details of Ablation Study can be found in Appendix C.5.
% \subsection{Discussion on Limitations}

% \noindent
% \textbf{Discussion on Limitations}\hspace{0.6mm}

\section{Conclusion}
% conclusion here.
We introduced FlexHB, a new HPO algorithm designed for efficient configuration searching and flexible resource allocation.
It utilizes fine-grained fidelity information, 
as well as going beyond the paradigm of Successive Halving and HyperBand. 
Experiment results on various HPO tasks demonstrate that FlexHB achieves 
prevailing efficiency compared to other competitive methods, including BOHB 
and MFES-HB. 
We also investigated effects of core components in our method, 
i.e., Fine-Grained Fidelity, GloSH, 
and FlexBand, showing their benefits for hyperparameter tuning.

\bibliography{aaai24}

\newpage
\clearpage
\appendix
% \section{Appendix}
\section{Appendix}
In Section A of Appendix, we show detailed pseudo-code for Successive Halving and 
HyperBand. In Section B, we provide details of the proposed FlexHB method and a 
brief discussion on its limitations. 
Section C contains full results and details of our experiments with FlexHB and other HPO methods, 
empirical results on LCBench are also included. 
\section{A. Successive Halving and HyperBand}
\label{appendix: SH and HyperBand}
Instead of evaluating every candidate with full resources, some configurations in Successive Halving(SH) are chosen 
to be ``early-stopped'', and only a small fraction of candidates can exploit higher-cost resources. 
Resources used in Successive Halving (i.e., budget) could be defined by running time, epochs of training, 
number of data samples and so on. 
% Procedure of SH is stated in Section \ref{subsec:SH}, and 
The pseudo-code is shown in Algorithm \ref{SH Algo}. 
 $Rank\_By\_Metric$ will rank candidate configurations by their evaluation metrics of current budget, and $Top\_K$ will select $K$ configurations with best metrics.
The value of $\eta$ is usually set to $3$.
\begin{algorithm}[hb!]
  % \SetKwInOut{Input}{input}\SetKwInOut{Output}{output}
   \begin{algorithmic}
    \STATE {\bfseries Input:} initial budget $b_0$, maximum budget $b_{max}$, set of $n$ configurations $C={c_1, ..., c_n}$
    \STATE $i = 0$ 
    \STATE $b_i = b_0, n_i = |C|$ 
    \WHILE{$b_i \leq b_{max}$}
    \STATE   Train and evaluate $C$ with $b_i$ resources
    \STATE $C_{i} = Rank\_By\_Metric(C_{i})$ 
    \STATE $C = Top\_K(C_{i}, \floor{n_i/\eta})$ 
    \STATE $b_i = b_i\eta$ 
    \STATE $n_i = \floor{n_i/\eta}$ 
    \STATE $i += 1$ 
    \ENDWHILE
    \end{algorithmic}
  \caption{Pseudo-code for Successive Halving(SH) used by HyperBand}
  \label{SH Algo}
\end{algorithm}
% \vspace{-4mm}

% As stated in Section \ref{subsec:HyperBand}, 
HyperBand(HB) address the ``$n$ vs $B/n$'' tradeoff by calling SH with different initialization. The pseudo-code for HB is shown in Algorithm \ref{HyperBand Algo}. HB keeps calling Successive Halving for multiple times. The value of $\eta$ is usually set to $3$.
\begin{algorithm}[h]
  % \SetKwInOut{Input}{input}\SetKwInOut{Output}{output}
 \begin{algorithmic}
   \STATE {\bfseries Input:} maximum resource budget $R$ that can be allocated for a configuration, hyperparameter space $\Omega$, and $\eta$
   \STATE {\bfseries Output:} best configuration with top evaluation performance
   \STATE $s_{max} = \floor{log_{\eta}{R}}, B = (s_{max} + 1)$
    \FOR{$s \in \{s_{max}, s_{max} - 1, 0\}$}
        \STATE $ n = \ceil{\frac{B}{R}\frac{\eta^s}{s+1} }, r=R\eta^{-s} $
        \STATE sample $n$ configurations from $\Omega$ as set $C$
        \STATE run SH with $n$, $r, R$ and $C$ as input
    \ENDFOR
 \end{algorithmic}
  \caption{Pseudo-code for HyperBand(HB)}
  \label{HyperBand Algo}
\end{algorithm}
\section{B. More details of FlexHB}
\subsection{B.1 Comparison of vanilla Multi-Fidelity and Fine-Grained Fidelity}
\label{appendix: MF vs FGF}
\begin{figure}[H]
    % \vspace{-3mm}
    \centering
    \includegraphics[width=0.9\columnwidth]{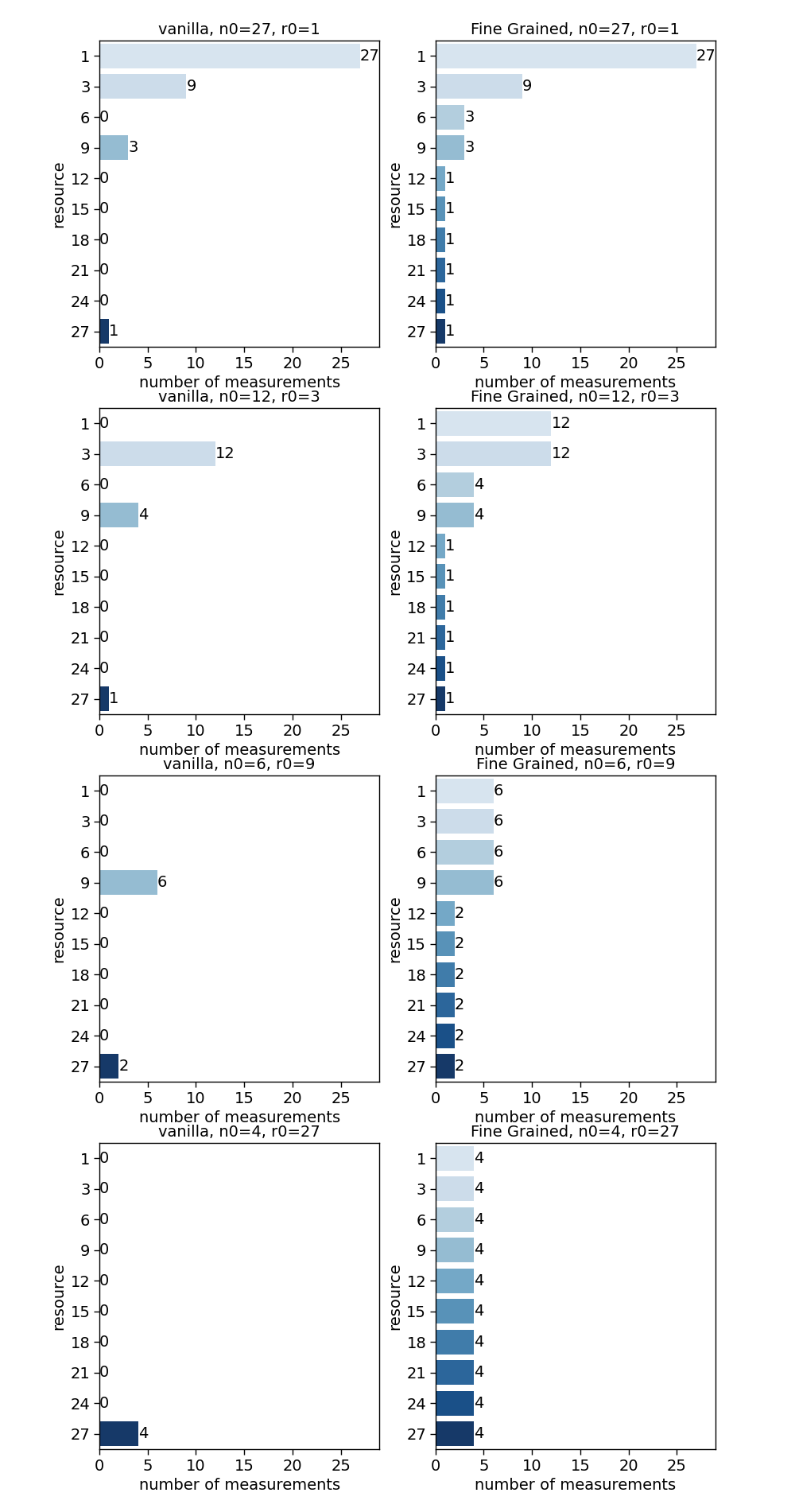}
    \caption{Comparison of measurement distribution over fidelity levels in vanilla Multi-Fidelity Method and FGF. Left column for vanilla and right column for FGF. Four brackets given $R=27$ are displayed row by row, the top row for the most exploring bracket and bottom row for the most exploiting bracket. Darker color indicates measurements for higher fidelity.}
    \label{fig:fgf_measurements_full_comparison}
\end{figure}

% In Section \ref{subsec: Fine-Grained Fidelity}, 
In this paper, 
we have shown the difference between measurements collected by vanilla multi-fidelity methods and Fine-Grained Fidelity (Figure \ref{fig:Fine Grained Illustration}). 
To fully demonstrate the superiority of FGF, we display a comprehensive comparison of measurement distributions on fidelity levels in Figure \ref{fig:fgf_measurements_full_comparison}. 
Fine-Grained Fidelity method allows the HPO algorithm to collect more measurements 
under every SH brackets, and alleviates the imbalance between information for 
lower fidelities and for higher fidelities.

\begin{figure}
    % \centering
    \begin{subfigure}[b]{0.24\columnwidth}
        \includegraphics[width=\columnwidth]{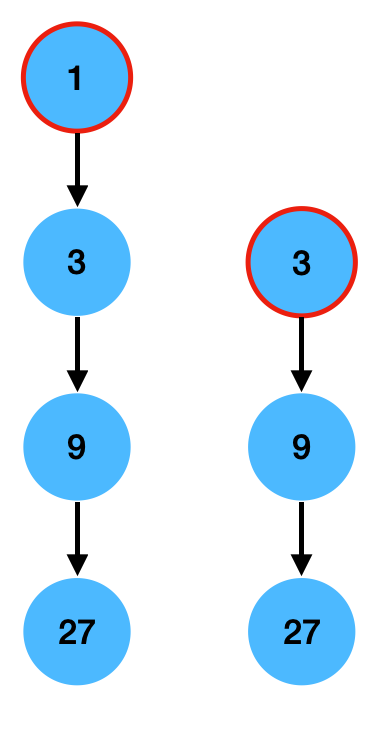}
        \caption{HyperBand}
        \label{}
    \end{subfigure}
    \begin{subfigure}[b]{0.24\columnwidth}
        \includegraphics[width=\columnwidth]{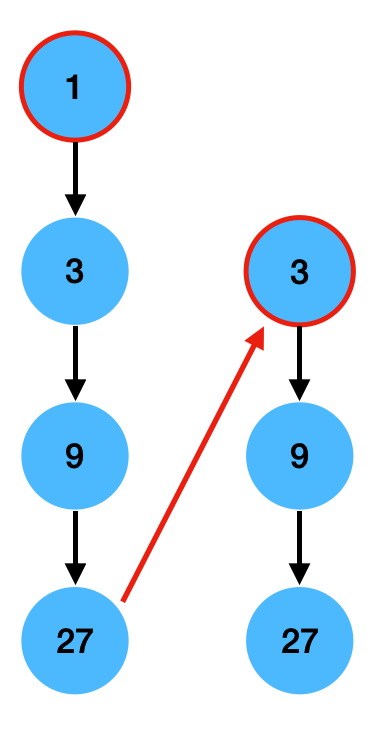}
        \caption{BOHB}
        \label{}
    \end{subfigure}
    \hfill
    \begin{subfigure}[b]{0.24\columnwidth}
        \includegraphics[width=\columnwidth]{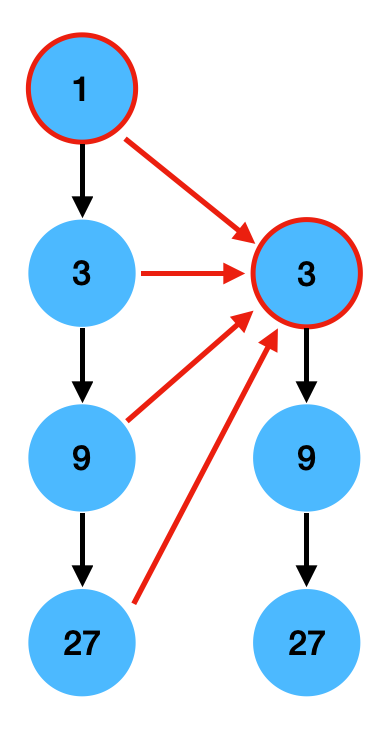}
        \caption{MFES-HB}
        \label{}
    \end{subfigure}
    \begin{subfigure}[b]{0.25\columnwidth}
        \includegraphics[width=\columnwidth]{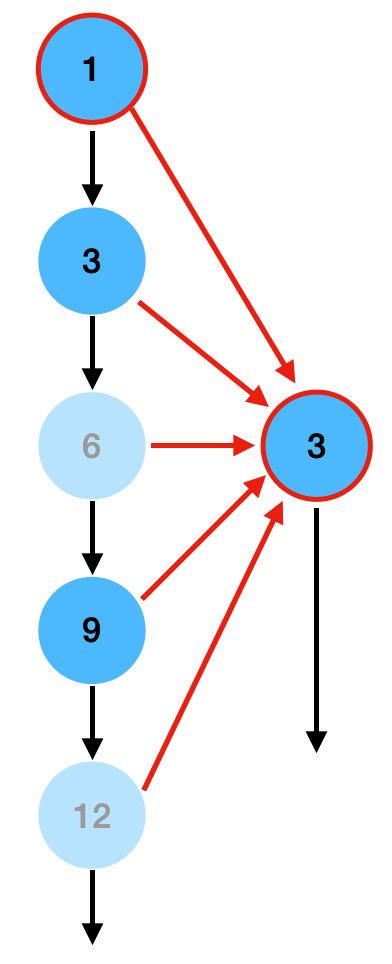}
        \caption{FlexHB}
        \label{}
    \end{subfigure}
    \caption{The generation of new configurations in different HPO methods. 
    For simplicity, only first two brackets in HyperBand are visualized. 
    Black arrow indicates the incremental training process, 
    number in blue cells indicates resource number for different fidelity levels, 
    lighter blue cells indicate fidelity levels NOT for early stopping selection in 
    Successive Halving. 
    Red-bordered cells are where new configurations are sampled. 
    Red arrow indicates the utilization of history measurements.}
    \label{fig:circles}
\end{figure}

Figure \ref{fig:circles} shows 
the design for generating new configurations in different HPO methods. 
HyperBand just randomly samples new candidates, 
BOHB utilizes full fidelity information from history records, 
and MFES-HB take advantage of multi-fidelity measurements from Successive Halving 
process. 
FlexHB is not constrained by Successive Halving, 
it uniformly collects measurements during the whole training process of a configuration.

\subsection{B.2 Weights of Surrogates in FGF}
The calculation of surrogate weights in FGF is similar to MFES-HB, 
except for the last (and usually the most important) surrogate $M_K$ fitted 
on full-fidelity measurements. To obtain a reasonable weight for $M_K$, 
we are not able to use the ``ranking loss'' 
mentioned in MFES-HB \cite{li2021mfes} 
because $M_K$ is trained on $D_K$ directly and we need a 
better estimate for the generalization ability of $M_K$. 
MFES-HB addresses this problem by introducing the cross-validation. 
First, $|D_K|$ leave-one-out surrogates $M_K^{-i}$ are trained on dataset $D_K^{-i}$, 
which is obtained by removing $i$-th measurement in $D_K$. Then the cross-validation ``ranking loss'' $L^{\prime}$ for $M_K$ is defined as:
\begin{equation}
L^{\prime}(M_K)=\sum_{j=1}^{|D_K|}\sum_{k=1}^{|D_K|}\textbf{1}((\mu_K^{-j}(X_j) < \mu_K^{-j}(X_k))\oplus(y_j < y_k))
\end{equation}
where $\mu_K^{-j}$ represents the prediction of $M_K^{-j}$. Finally, the weights is calculated based on percentage $p_i$ of consistent ordered pairs, that is:
\begin{equation}
\label{eq: pi calculation}
    p_i = 1 - \frac{L(M_i)}{\#\ of\ pairs\ in\ D_K}
\end{equation}
\begin{equation}
    \label{eq: wi calculated}
    w_i = \frac{p_i^\gamma}{\sum_{k=1}^{K}{p_k^\gamma}}
\end{equation}

However, this formulation of weight $w_K$ is inappropriate in our case, because:\\
(1) $L^{\prime}(M_K)$ is not consistent with $L(M_1), ..., L(M_{K-1})$ in concept. 
Applying Formula \eqref{eq: pi calculation} directly with $L^{\prime}(M_K)$ will 
lead to different implications of $p_K$, compared with other $p_1, ..., p_{K-1}$.\\
(2) empirical statistics show that the $w_K$ calculated as above can be quite smaller than other $w_i$ (see Figure \ref{fig:weights_mfse}).\\
(3) FlexHB builds surrogates with linearly increasing fidelities, and 
% there are more surrogates in FGF than in traditional method, e.g., 
% we have to obtain $w_1, ..., w_9$ instead of $w_1, ..., w_4$ while $\eta=3, R=27$. 
% Thus, 
the surrogate $M_{K-1}$ shall have similar weight with $M_K$.

Consequently, in FlexHB, we propose a new formulation of weight $w_K$ for $M_K$:\\
(1) obtain the cross-validation ranking loss $L^{\prime}(M_{K-1})$ and $L^{\prime}(M_{K})$, note that they are often fitted on same number of data points ($|D_{K-1}| = |D_K|$).\\
(2) calculate $p_{K-1}$, 
where $p_{K-1} = 1 - \frac{L(M_{K-1})}{\#\ of\ pairs\ in\ D_K}$, 
and $L(M_{K-1})$ is the original ranking loss 
(Formula \eqref{equation: MFES ranking loss}) 
based on predictions of $M_{K-1}$ and measurements in $D_K$.\\
(3) get the \textit{simulated} percentage of consistent order pairs for $M_K$, namely $p_K = p_{K-1}\cdot\frac{L^{\prime}_K}{L^{\prime}_{K-1}}$. \\
(4) get $w_K$ according to Formula \eqref{eq: wi calculated}.

Intuitively, our new $w_K$ is proportional to the generalization ability of $M_K$ (by comparing $L^{\prime}_K$ and $L^{\prime}_{K-1}$), and the numerical difference between $w_{K-1}$ and $w_{K}$ will not be too large. Note that sometimes the simulated $p_K$ can be larger than $1$, and will be handled by $p_K = min(0.99, p_K)$.

To yield the integrated prediction of BO surrogates,
% given surrogate models,
we can employ the weight vector to do a weighted sum of predicted mean and variance 
from each base surrogates, or use the 
gPoE\cite{cao2014generalized} method as in \cite{li2021mfes}.

\label{appendix: weights of FGF}
\subsection{B.3 Details of GloSH}
\label{appendix: GloSH}
Detailed design of GloSH can be found in Algorithm \ref{Glosh Algo Main}. $Merge$ function will do a union for two sets, $Rank\_By\_Metric$ will sort configurations from best evaluation metric to worst, and $random$ is a random number (in $0$ to $1$) generator. Probability $\lambda_i$ is used to control the intensity of global comparison, and $\eta$ is set to $3$.
We use $\lambda_0=\frac{1}{3}$, $\lambda_1=\frac{1}{2}$, $\lambda_2=1$ while $r_0=1$, $r_1=3$, $r_2=9$ in a typical HyperBand setting with $R=27$. 
\begin{algorithm}[h]
  % \SetKwInOut{Input}{input}\SetKwInOut{Output}{output}
  \begin{algorithmic}      
   \STATE {\bfseries Input:} initial budget $b_0$, maximum budget $b_{max}$, 
  set of $n$ configurations $C = \{c_1, ..., c_n\}$, 
  probabilities for using the old stopped configurations $\{\lambda_0, ..., \lambda_m\}$,
  set of early-stopped configurations for each budget levels $S_0, ..., S_m$
  \STATE {\bfseries Output:} updated $\{S_0, ..., S_m\}$
   \STATE $i = 0$, $b_i = b_0, n_i = |C|$ 
   % $b_i = b_0, n_i = |C|$ \;
  \WHILE{$b_i \leq b_{max}$}
    \STATE Train and evaluate $C$ with $b_i$ resources 
    \STATE $C_{i,global} = Merge(C, S_i)$ 
    % \tcp*[h]{\smaller{merge current and old configurations into one set}} \;
    
    \STATE $C_{i,global} = Rank\_By\_Metric(C_{i,global})$ 
    % \tcp*[h]{\smaller{rank configurations in descending order by their performance}}\;
    
    \STATE Initialize $C_{keep}$, $C_{throw}$ as $\varnothing$ 
    \WHILE{$|C_{keep}| < \floor{n_i / \eta} $}
        \STATE $ c_j$ is the $j$-th element in $C_{i,global}$ 
        \IF{$c_j$ is from $S_i$}
            \IF{$random() < \lambda_i$}
                \STATE Add $c_j$ to $C_{keep}$ 
            \ENDIF
        \ENDIF
        \IF{$c_j$ is from $C$}
            \STATE Add $c_j$ to $C_{keep}$ 
        \ENDIF
        \STATE $j = j + 1$ 
    \ENDWHILE
    \STATE $C = C_{keep}$ 
    \STATE $C_{throw} = C_{i,global} - C_{keep}$ 
    \STATE $S_i = C_{throw}$ 
    \STATE $b_i = b_i \eta$, $n_i = \floor{n_i / \eta}$ 
    % $n_i = \floor{n_i / \eta}$ \;
    \STATE $i = i + 1$ 
  \ENDWHILE
\end{algorithmic}
  \caption{Pseudo-code for GloSH}
  \label{Glosh Algo}
\end{algorithm}

\subsection{B.4 Details of FlexBand}
\label{appendix: FlexBand}
Algorithm \ref{Algo:FlexBand Main} gives the pseudo-code for FlexBand. 
Firstly we generate the bracket arrangements as in HyperBand, 
then compute the rank correlation metric $\tau$ for each adjacent pairs of 
fidelity levels, 
and further modify the arrangement of brackets according to $\tau$. 
Eventually, FlexBand can call SH or GloSH following the new arrangement. 

\begin{figure}[h]
    \centering
    \includegraphics[width=0.92\columnwidth]{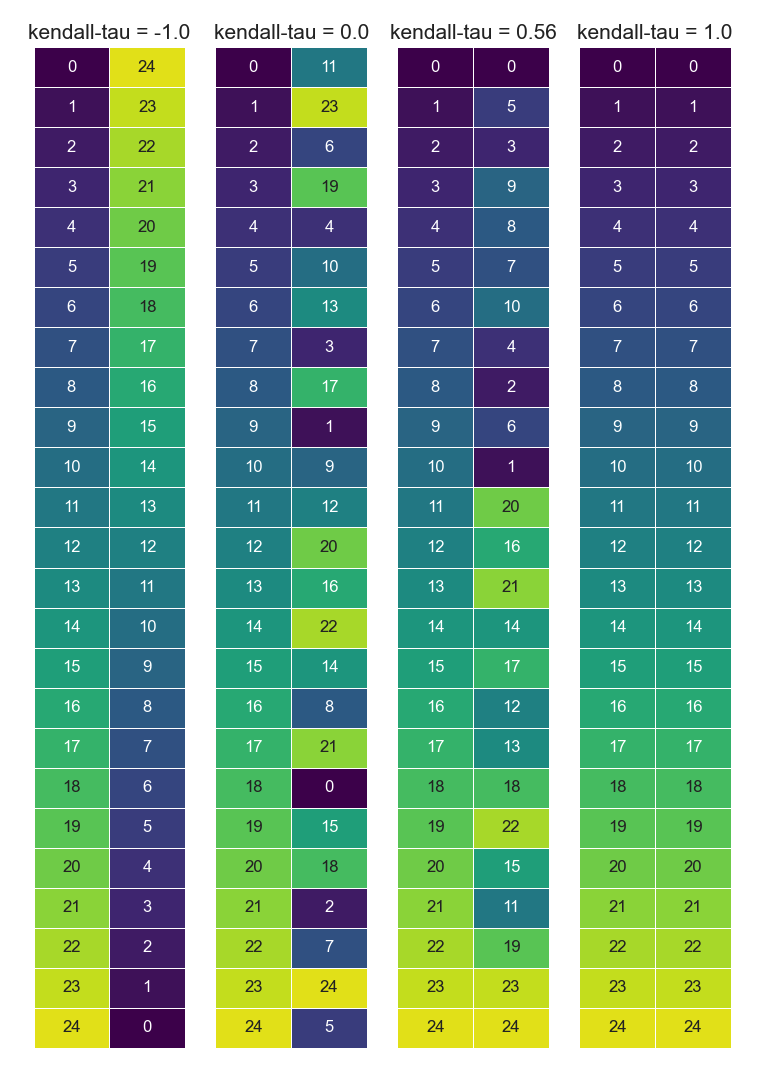}
    \caption{Visualizing rank correlation under four different Kendall-tau values. The length of sequence is $25$. Number in cells indicates the relative order within the array, darker colored cells for smaller values and lighter colored cells for bigger values. Color distribution of two lists becomes more consistent as Kendall-Tau value increases.}
    \label{fig:kendall_tau_visual}
    % \vspace{-1mm}
\end{figure}

The rank correlation statistics $K$ in pseudo-code is exactly Kendall-tau metric $\tau$,
 and $K_{thres}$ is set to $0.55$. 
Figure \ref{fig:kendall_tau_visual} visualizes pairs of two sequences and their 
corresponding Kendall-tau values, 
we can tell that the rank correlation when $\tau \geq 0.55$ is quite similar 
to the case where 
the ranking are identical ($\tau = 1.0$).

\begin{algorithm}[h]
\begin{algorithmic}
      % \SetKwInOut{Input}{input}\SetKwInOut{Output}{output}
  \STATE {\bfseries Input:} maximum resource budget $R$, hyperparameter space $\Omega$, history records dataset $D = \{D_{r_1}, ..., D_{r_m}$\} for each fidelity level $r$, the threshold $K_{thres}$, and $\eta$
 \STATE {\bfseries Output:} best configuration
  % with top evaluation performance
    \STATE Initialize $\textbf{N}, \textbf{R}$ as $\varnothing$ \;
    \STATE $s_{max} = \floor{log_{\eta}{R}}, B = (s_{max} + 1)$ \;
    % $m = 0$
    \FOR{$s \in \{s_{max}, s_{max} - 1, ..., 0\}$}
        \STATE $ n = \ceil{\frac{B}{R}\frac{\eta^s}{s+1} }, r=R\eta^{-s} $ \;
        \STATE Append $n$ into $\textbf{N}$, $r$ into $\textbf{R}$ \;
    \ENDFOR
    \STATE Get rank correlation $K$ for each pair of $D_{r_i}, D_{r_{i-1}}$ \;
    \STATE Make a copy of $\textbf{N}$,$\textbf{R}$ as $\textbf{N}^\prime,\textbf{R}^\prime $ \;
    \FOR{$j \in \{ |\textbf{N}|, |\textbf{N}| - 1, ..., 1\}$}
        \STATE $i  = j - 1$ \;
        \STATE $r_i, r_j, n_i, n_j$ are $i$-th and $j$-th element of $\textbf{R}$,$\textbf{N}$  \;
        \IF{$K_{r_i, r_j} > K_{thres}$}
              \STATE $\textbf{N}^\prime_{j} = n_i$, $\textbf{R}^\prime_{j} = r_i$  \;
               % $\textbf{R}^\prime_{j} = r_i$
        \ENDIF
    \ENDFOR
    \FOR{$i \in \{0, 1, ..., |\textbf{N}| \}$}
        % $r_i, n_i$ are $i$-th element of $\textbf{R}$,$\textbf{N}$  \;
        \STATE sample $n_i$ configurations from $\Omega$ as set $C$\;
        \STATE run SH or GloSH given $C$, $n_i$ and $r_i$\;
    \ENDFOR
\end{algorithmic}
  \caption{Pseudo-code for FlexBand}
  \label{Algo:FlexBand}
\end{algorithm}

In the early stage of tuning, there are only a few history records available and the obtained 
correlation metric can be flawed. To avoid this issue, 
% FlexBand will generate new arrangements only after enough history records are collected. 
we employ a warm-up number $h$ and the FlexBand will generate new arrangements 
only after number of history records for each fidelity level exceeds the warm-up 
number, i.e., $min{|D_{r_i}|} \geq h$. In our experiments, $h$ is set to $25$. 

\subsection{B.5 Hyperparameters of FlexHB}
\label{appendix: Hyperparameters of FlexHB}
In Table \ref{tab:hp of FlexHB}, we summarize hyperparameters in our method and recommended values we use in experiments. Note that only additional hyperparameters compared to BOHB are listed (e.g., $\eta$ is not included.).
% \usepackage{graphicx}
% \vspace{-3mm}
\begin{table}[]
% \vskip 0.1in
\centering
\resizebox{\linewidth}{!}{
\begin{tabular}{l|l|l}
\hline
Name &
  Usage &
  Recommended Settings \\
  \hline
% $\eta$ &
%   same as eta in HyperBand &
%   $\eta = 3$ \\
  % \hline
$g$ &
  \begin{tabular}[c]{@{}l@{}}mininum gap of resources for\\ fine-grained measurements\end{tabular} &
  $g=\eta=3$ \\
  \hline
$\lambda$ &
  \begin{tabular}[c]{@{}l@{}}probability for using old terminated\\ configurations in global ranking\end{tabular} &
  \begin{tabular}[c]{@{}l@{}}$\lambda$ shall increase as fidelity level grows.\\ with $R=27$, for $r0=1,..., r3=27$,\\ we set $\lambda_0 = 1/3, \lambda_1 = 1/2, \lambda_2 = 1$.\\ with $R=81$, for $r0=1,..., r4=81$\\ we set $\lambda_0 = 1/4, \lambda_1 = 1/3, \lambda_2=1/2, \lambda_3=1$.\end{tabular} \\
  \hline
$K_{thres}$ &
  \begin{tabular}[c]{@{}l@{}}threshold for Kendall $\tau$\\ between adjacent fidelity levels\end{tabular} &
  $K_{thres} = 0.55$ \\
   \hline
  $h$ & \begin{tabular}[c]{@{}l@{}}warm-up number of history records for\\  adjusting brackets in FlexBand.\end{tabular}
   & $h=25$ \\
  \hline
% $p_r$ &
%   \begin{tabular}[c]{@{}l@{}}probability to use randomly \\ sampled configurations\end{tabular} &
%   \begin{tabular}[c]{@{}l@{}}$p_r = 0.2$\\ 
%   \end{tabular} \\ \hline
% \hline
\end{tabular}
% \hline
}
\caption{Hyperparameters in FlexHB}
\label{tab:hp of FlexHB}
% \vskip -0.1in 
\end{table}

\subsection{B.6 Theoretical Discussion}
\label{appendix:theory}
% ! TODO
\textit{Note that we list referenced papers at the end of this section.}

\textbf{1. Theoretical guarantee of using FGF in SH}

Following Section 5.4 of [1], we model HPO as a NIAB problem with finite horizon setting. The key theoretical conclusions of SH in [1] is that:

(1) If SH is run with any budget $ B \geq  Z_{SH}$ then an arm $\widehat{i}$ is returned that satisfies $v_{\widehat{i}} - v_1 \leq \epsilon/2,$ where $Z_{SH}$ is determined by $\eta, R, n$ (Theorem 8 of [1]). Also note that $v_i = \lim_{\tau\to\infty} l_{i, \tau}$ and $v_1 \leq ... \leq v_{n}$.

(2) Fix $\delta \in (0,1)$ 
and $\epsilon \geq 4(F^{-1}(\frac{log(2/\delta)}{n}-v_*))$. 
Let $H$ be as defined in Lemma 2 and $B=\eta log_{\eta}(R)(n+max(R, H))$. 
If SH is run with the specified $B$ and $n$ configurations drawn randomly with $F$, 
then an arm $\widehat{i} \in [n] $ is returned s.t. with prob at least $1-\delta$ we have 
$v_{\widehat{i}}-v_{\*}\leq (F^{-1}(\frac{log(2/\delta)}{n})-v_{*})+\epsilon/2$. (Corollary 9 of [1]).

Note that FGF has no effect on $F, \gamma, n$. 
As for $R$, the additional cost is for testing (not training) confs on 
non-earlystopping levels (e.g., 
when $\eta=3$, $r=6$ is a non-earlystopping level and $r = 9$ is not). 
In actual HPO for ML models, cost for testing is often negligible compared to training (e.g., training a NN needs tons of epochs on train set while testing needs only one iteration of test set, which is much cheaper). Also, the total additional times of testing is $\mathcal{O}(R)$. Thus the total additional cost, as a product of test cost and test times, is negligible to the original $R$. 

Specifically, the proof for Thm.8 in [1] (page 48) utilizes the fact that 
$ \sum_{k=0}^{s}n_kr_k \leq nR(s+1)\eta^{-s} \leq B$, which still holds true in FGF. 
Further, as FGF has no effects on $v_i$, $max(\frac{\epsilon}{4}, \frac{v_i - v_1}{2})$ and $|S_k|$, the derivation of the relation between $r_k$ and $min(R, \gamma^{-1}(max(\frac{\epsilon}{4}, \frac{ v_{\lfloor{|S_k|/2 \rfloor{}}+1}-v_{1} }{2})))$ still holds, as well as various cases discussed in the proof for Thm.1 and Lemma 2 (note that Thm.8 and Cor.9 are the finite horizon version of Thm.1 and Lemma 2). 
Thus, Thm.8 and Cor.9 are also applicable for FGF.  

\textbf{2. Analysis of GloSH}.

Since Thm.8 and Cor.9 for SH are the finite horizon version of Thm.1 and Cor.2 in [1], we will refer to the proof in Appendix B.1, B.2 and B.7 of [1].

Recalling that GloSH will rank all confs (both current running and old killed ones) at any round $k$, from the top arm (i.e., the conf with lowest validation error $l_k$), GloSH picks one by one till number of picked arms meets $n_k/\eta$. For old terminated ones, GloSH will pick it with prob $\lambda_r \in (0, 1)$ (ref to Algo.3 
% in our Appendix
).

For Thm.8 (and Thm.1), firstly, it can be easily shown that the bound for total resources still holds, i.e., $\sum_{k=0}^{\lceil{\log_2(n)\rceil-1}}|S_k|\lfloor \frac{B}{|S_k|\lceil log(n) \rceil} \rfloor \leq B$. And, though some old terminated conf may 
``replace'' current confs, the property of $\gamma$ (Eq. 8 of [1]) still holds.

Considering a bracket, for the ending round $K$ where it will 
return $|S_K|$ arms as the final choices, we now examine the following 
exhaustive cases, we denote $v_{c,i,k}$ for a current conf (arm $i_c$) and 
$v_{o,i,k}$ for a old ``global'' conf $i_o$. 
In the finite horizon version, 
naturally we have $v_i = v_{i,K}$ for both $i_c$ and $i_o$ (as $r_K$ being the 
maximal $R$). 
As in [1], we assume $v_{1} \leq v_{2} … \leq v_{n}$, where $n := |S_0|$.

Case 1: $\lambda_k = 0$. In this case, GloSH degrades to SH.

Case 2: $\lambda_k > 0$, and $S_K \subseteq S_0$. Then, GloSH actually degrades to SH.

Case 3: $\lambda_k > 0$, and $S_K \cap S_0 \neq \emptyset$. 
Recalling that at any round $k$ the GloSH algorithm will eliminate an arm $i_c$ 
with $v_{c,i,k} < v_{c, \lfloor{|S_k|/2 \rfloor{}} + 1, k}$ only 
if there exists an arm $i_o$ with $v_{o,i,k} < v_{c,i,k}$. 
Thus, for any $i_c \in S_K \cap S_0$, at any preceding round $k$ we 
have $v_{c,i,k} < v_{c, \lfloor{|S_k|/2 \rfloor{}} + 1, k}$. 
This implies the ``purity'' of returned arms in GloSH, 
indicating that inferior confs sampled in $S_0$ will not remain in $S_K$. 
Therefore, analogous conclusions like Thm.8 also hold for GloSH, 
by only considering those $i_c$ that lies in $S_K \cap S_0$ for each round $k$ in the proof. With some $i_c$ crowded out by $i_o$ at each round, 
Case 3 is equivalent to SH with a smaller $S_0$ (or larger $\eta$).

Case 4: $\lambda_k > 0$, and $S_K \cap S_0 = \emptyset$. Then we have $i_c \in S_0$ and $i_c \notin S_K$ (by definition).
This can be intepreted as a ``larger'' SH with a new $S_0^{\prime}$. 
$S_0^{\prime} = S_0 \cup S_g$ where 
$S_g = \{i_{o,k} | v_{o,i,k} < v_{c,i,k}, v_{c,i,k} < 
v_{c, \lfloor{|S_k|/2 \rfloor{}} + 1, k}\}$ for each $k$, 
and $v_{o, \lfloor{|S_k|/2 \rfloor{}} + 1, K} < v_{c, 1, K}$. 
In this case, Thm.8 still holds for the ``larger'' SH with $S_0^\prime$, 
indicating that an arm $\widehat{i}$ is returned that satisfies 
$v_{\widehat{i} - v_1} \leq \epsilon/2$ 
for $\widehat{i} \in i_c \cup i_o$, $v_1 = min(v_{c,i} \cup v_{o, i})$. 
It is also worth mentioning that the guarantee does not hold now for the original 
$v_1, ..., v_n$ for all $i_c \in S_0$, since there exists a round $k$ where 
$S_k \cap S_0 = \emptyset$. After this round $k$, 
the GloSH is actually conducting SH based on another initial set of arms instead of 
$v_1, ..., v_n$ and $S_0$.

Similar conclusion can be established for Cor.9 following the inference above.

\textbf{3. Analysis of FlexBand}

The FlexBand is designed based on the finite horizon HB beacause only in such case the number of brackets in an HB iteration "is fixed at $\log_\eta
(R) + 1$"(Page 30 of [1]). And the theoretical analysis of [1] (which is based on infinite horizon HB) is not applicable for FlexBand since the number of unique brackets in infinite horizon version grows over time (ref to Section 3.5 of [1]), while FlexBand requires constant number of unique brackets to apply the flexible substitution.

However, it's worth noting that  "in scenarios where a max resource is known, it is better to use the finite horizon algorithm" (Section 4.1 of [1]), which is exactly the case for ML model tuning. 
And the modified bracket assignment in FlexBand could be interpreted as a "shifted" or "cutted" version of original HB, which is sanctioned within the proposed paradigm of [1].
As stated in Section 3.6 of [1]:
"if overhead is not a concern and aggressive exploration is desired, one can ... still use $\eta = 3$ or $4$ but only try brackets that do a baseline level of exploration, i.e., set $n_{min}$ and only try brackets from $s_{max}$ to $s=\lfloor log_{\eta}(n_{min}) \rfloor$."

In practice, SH or (finite version of) HB are all executed for multiple times till a satisfying conf is found (while theoretical analysis are all established for one iteration). And the overall efficiency of HPO is determined not only by evaluation scheme, but also the sampling strategy, which directly leads to researches like BOHB[2] and DEHB[3], the latter turns SH into a Differential Evolution process.

\textit{Reference:}

\textit{[1] Hyperband: A Novel Bandit-Based Approach to Hyperparameter Optimization}

\textit{[2] BOHB: Robust and Efficient Hyperparameter Optimization at Scale}

\textit{[3] DEHB: Evolutionary Hyperband for Scalable, Robust and Efficient Hyperparameter Optimization}

\subsection{B.7 Discussion on Limitations}
FGF requires collecting measurements on resource levels that are not for early-stopping, 
% as mentioned before, the consequent additional cost for testing confs is acceptable for most cases. 
yet there exists some problem settings where collecting intermediate measurements can be expensive, 
e.g., a ML model unable to save its trained state because 
of the particular implementation. 
Another limitation we observed is the sensitivity to the choice of probability factor 
$\lambda_i$ in GloSH that controls whether or not to pick an old stopped candidate. 
In practice, we intuitively assign larger 
$\lambda$ value for higher fidelity level. 
However, it is not clear a-priori how to determine best values for $\lambda_i$, 
and that can be an interesting avenue for future work. 
Lastly, a meta-learning (or transfer learning) paradigm could be applied for addressing the “$n$ vs $B/n$ trade-off”, 
which may outperform the heuristic design of HyperBand and FlexBand.

\section{C. Details of Experiments}
% \subsection{Details of HPO tasks}
\subsection{C.1 Experiment Settings}
% \label{appendix: More details of HPO task settings}
\begin{table}[h]
% \vskip 0.1in
\centering
\resizebox{\columnwidth}{!}{%
\begin{tabular}{|l|l|l|c|c|c|c|}
\hline
Task Type &
  ML Model &
  Dataset &
  $|\Omega|$ &
  \begin{tabular}[c]{@{}l@{}}Time\\ Limit\end{tabular} &
  \begin{tabular}[c]{@{}l@{}}Resource\\ Unit\end{tabular} &
  R \\ \hline
\multirow{3}{*}{Neural Networks} &
  MLP &
  MNIST &
  9 &
  5 hrs &
  1 epoch &
  27 \\ \cline{2-7} 
 & ResNet      & CIFAR-10      & 6 & 24 hrs & 0.5 epoch    & 81 \\ \cline{2-7} 
 & LSTM & Penn Treebank & 10 & 12 hrs & 0.5 epoch     &  27  \\ \hline
\multirow{5}{*}{XGBoost} &
  \multirow{5}{*}{XGBoost} & Covtype & 10 & 7.5 hrs & 1/27 dataset & 27 \\ \cline{3-7} 
 &             & Pokerhand     & 10 & 3 hrs & 1/27 dataset &27  \\ \cline{3-7} 
 &             & Adult         & 10 & 8000 secs &1/27 dataset &27 \\ \cline{3-7} 
 &             & Bank          & 10 & 10000 secs &1/27 dataset &27  \\ \hline
\end{tabular}%
}
\caption{HPO tasks settings. $|\Omega|$ is the dimension of hyperparameter space, ``Time Limit'' is for the whole tuning process.}
\label{tab:HPO tasks setting}
% \vskip -0.15in
\end{table}
\label{appendix: Hyperparameter Space}
We summarize experiment settings in Table \ref{tab:HPO tasks setting}. The detailed search space for each HPO tasks are as below:\\
(1) MLP: Search space $\Omega$ includes learning rate, learning rate decay, momentum, number of neuron units for two layers, batch size, dropout rates for layer outputs and the weight decay (L2 regularizer). Total number of hyperparameters to optimize is $9$.\\
(2) ResNet: Search space $\Omega$ includes batch size, learning rate, learning rate decay, momentum, weight decay (L2 regularizer) and an option that whether to use nesterov momentum. Total number of hyperparameters to optimize is $6$.\\
(3) LSTM: Search space $\Omega$ includes learning rate, batch size, embedding size, hidden size, dropout rates, optimizer choices, weight decay (L2 regularizer), range of initial values for network parameters, value for clipping gradients. Total number of hyperparameters to optimize is $10$.\\
(4) XGBoost tasks: Search space $\Omega$ includes eta(also known as learning rate), gamma, alpha (L1 regularization of weights), lambda (L2 regularization of weights), maximum depth of trees and some other hyperparameters used in official XGBoost implementation. Total number of hyperparameters to optimize is $10$.\\

For the MLP task, no further data augmentation is applied to images. We randomly split 20\% of training set as the validation set. For the ResNet task, We follow the classic settings, i.e., 40000 images as train set, 10000 images as validation set. Random cropping and horizontal flips are applied as data augmentation. For the LSTM task, we follow the original split of training and test(validation) set of Penn Treebank. In XGBoost experiments on Covertype, Adult and Bank, we randomly sample 80\% of whole dataset for training, 20\% for validation set. As for Pokerhand, we follow the original split of training and testing(validation) set.

For Neural Network tasks, the FGF gap $g$ is equal to $\eta$. For XGBoost tasks, since we use small subsets of training set as the resource unit, we drop most of fine-grained fidelity levels to cut unnecessary time costs, and use $r=1,6$ only. For example, given a SH bracket with $r0=3, R=27$, our fine fidelity levels in FlexHB will be $r=1,3,6,9,27$, here $r=3,9$ are fidelity levels for early stopping.

For HyperBand-like algorithms including BOHB, MFES-HB and our new FlexHB, $p_r = 0.2$ is applied, meaning that these methods sample constant fraction (20\%) of configurations randomly. If $p_r$ is set to 1.0, MFES-HB and BOHB will degrade to HyperBand, and FlexHB will degrade to a combination of random sampling strategy and FlexBand(with GloSH).

\subsection{C.2 Detailed Results of Experiments}
\label{appendix: full experiments}
\begin{figure*}[h]
    % \centering
    \begin{subfigure}[b]{0.33\linewidth}
        \includegraphics[width=\linewidth]{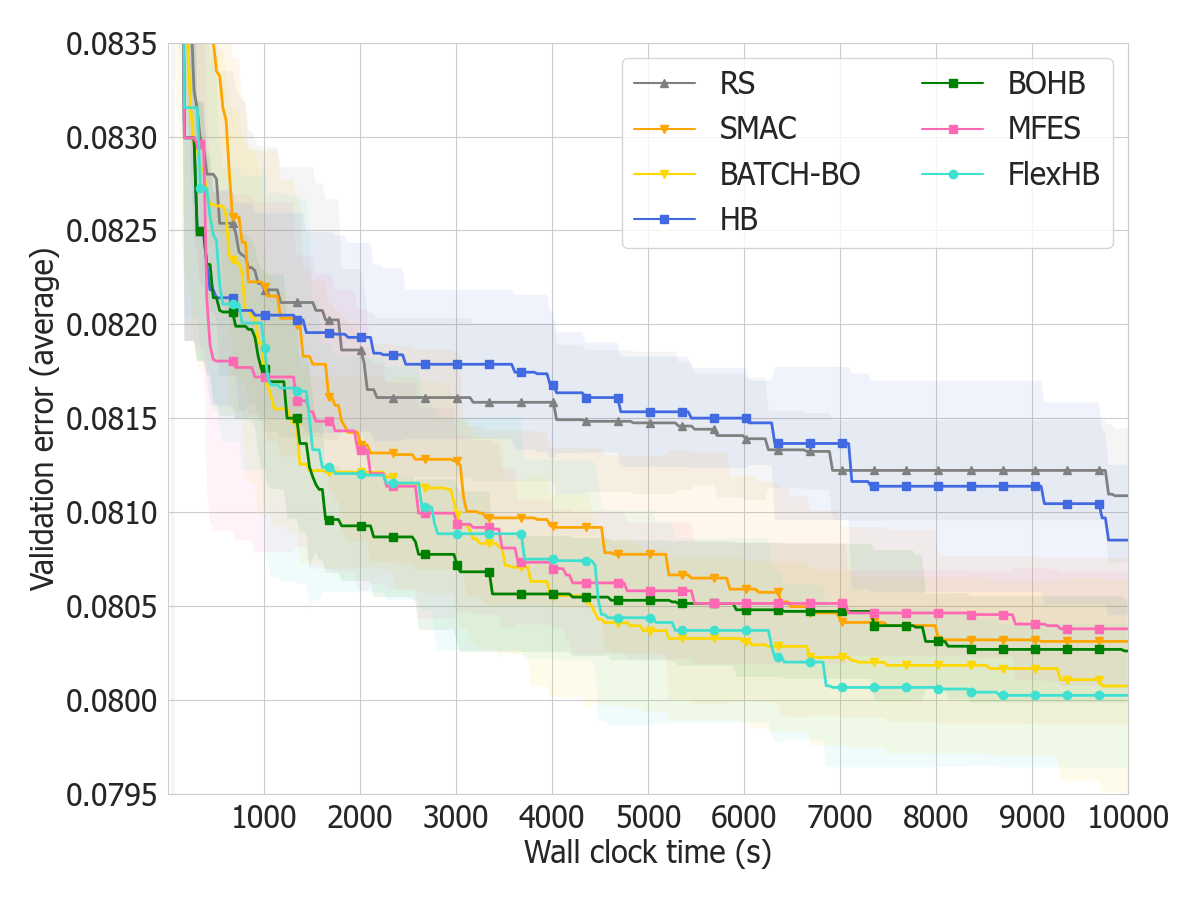}
        \caption{Bank}
        \label{MainExp_bank}
    \end{subfigure}%
    \begin{subfigure}[b]{0.33\linewidth}
        \includegraphics[width=\linewidth]{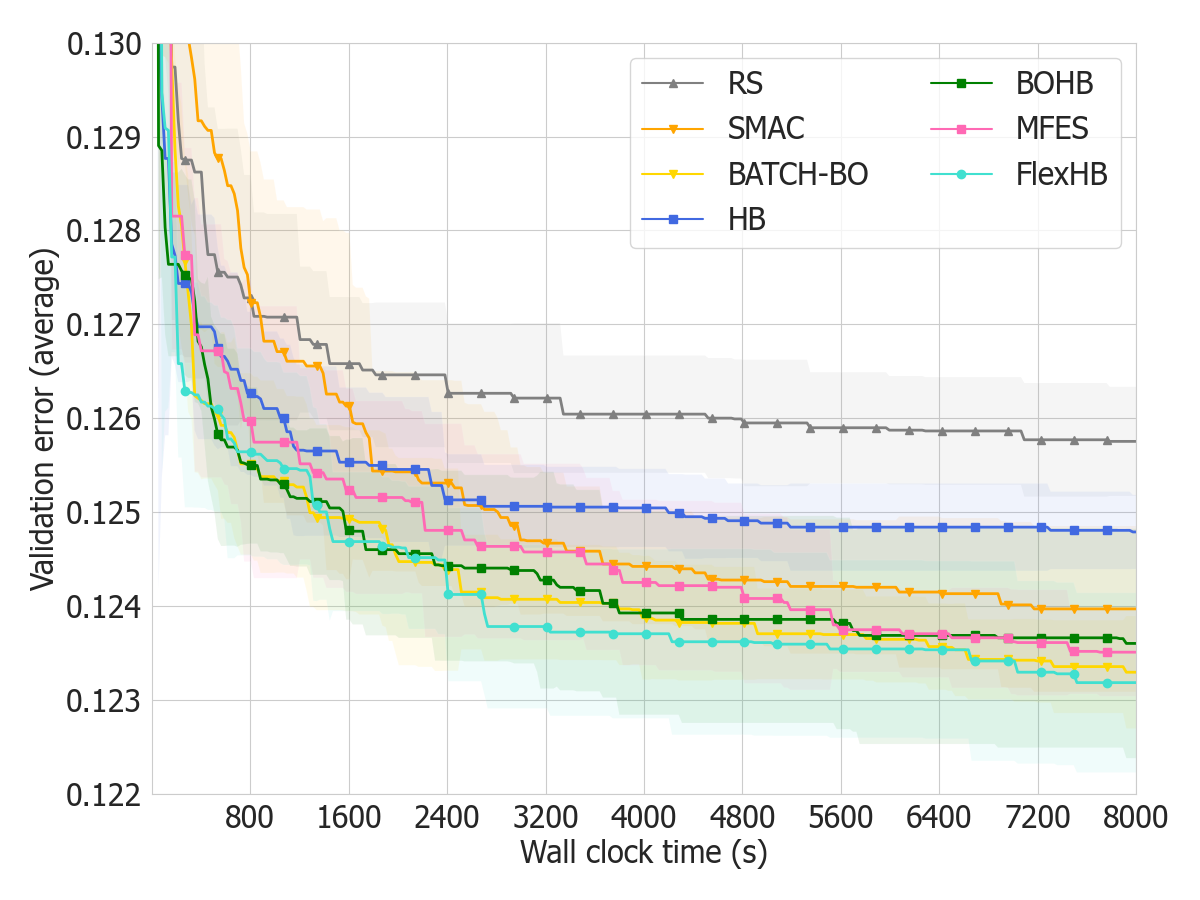}
        \caption{Adult}
        \label{MainExp_income}
    \end{subfigure}%
    \begin{subfigure}[b]{0.33\linewidth}
        \includegraphics[width=\linewidth]{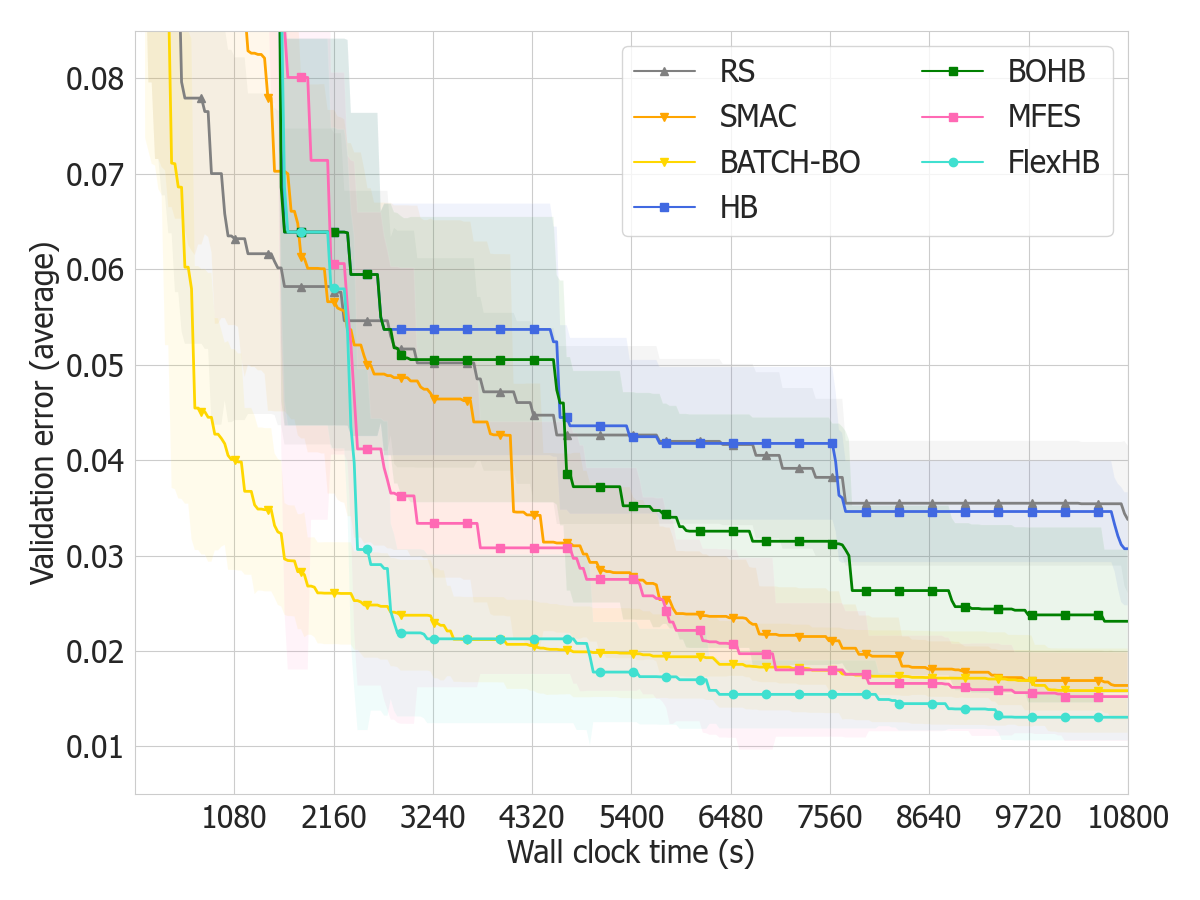}
        \caption{Pokerhand}
        \label{MainExp_pokerhand}
    \end{subfigure}
    \caption{Results for XGBoost Tasks}
    \label{fig:XGBoost Full Results}
    % \vspace{-3mm}
\end{figure*}
\begin{figure*}[t!]
    % \vspace{-3mm}
     \centering
     \begin{subfigure}[b]{0.33\linewidth}
         \centering
         \includegraphics[width=\linewidth]{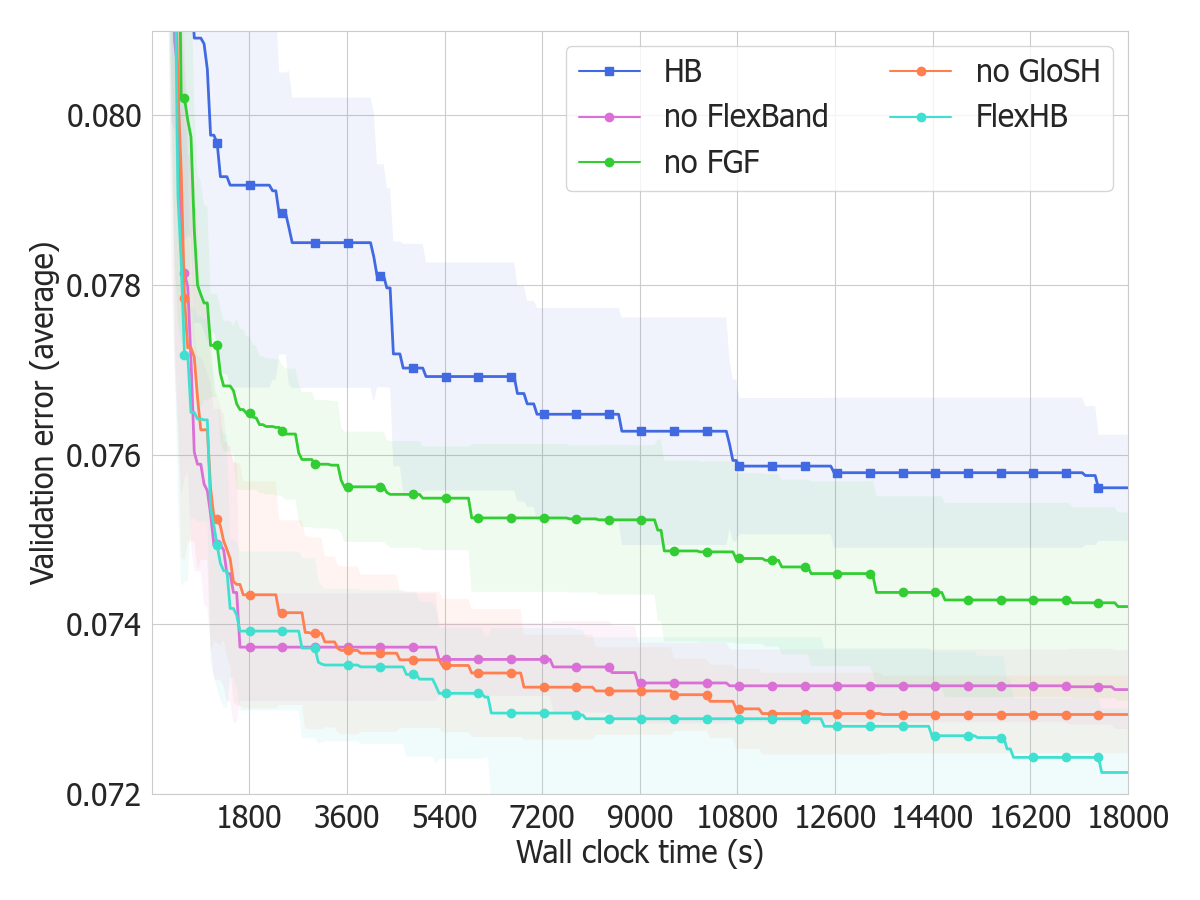}
         \caption{MLP on MNIST}
         \label{fig:abl MNIST}
     \end{subfigure}
     \hfill
     \begin{subfigure}[b]{0.33\linewidth}
         \centering
         \includegraphics[width=\linewidth]{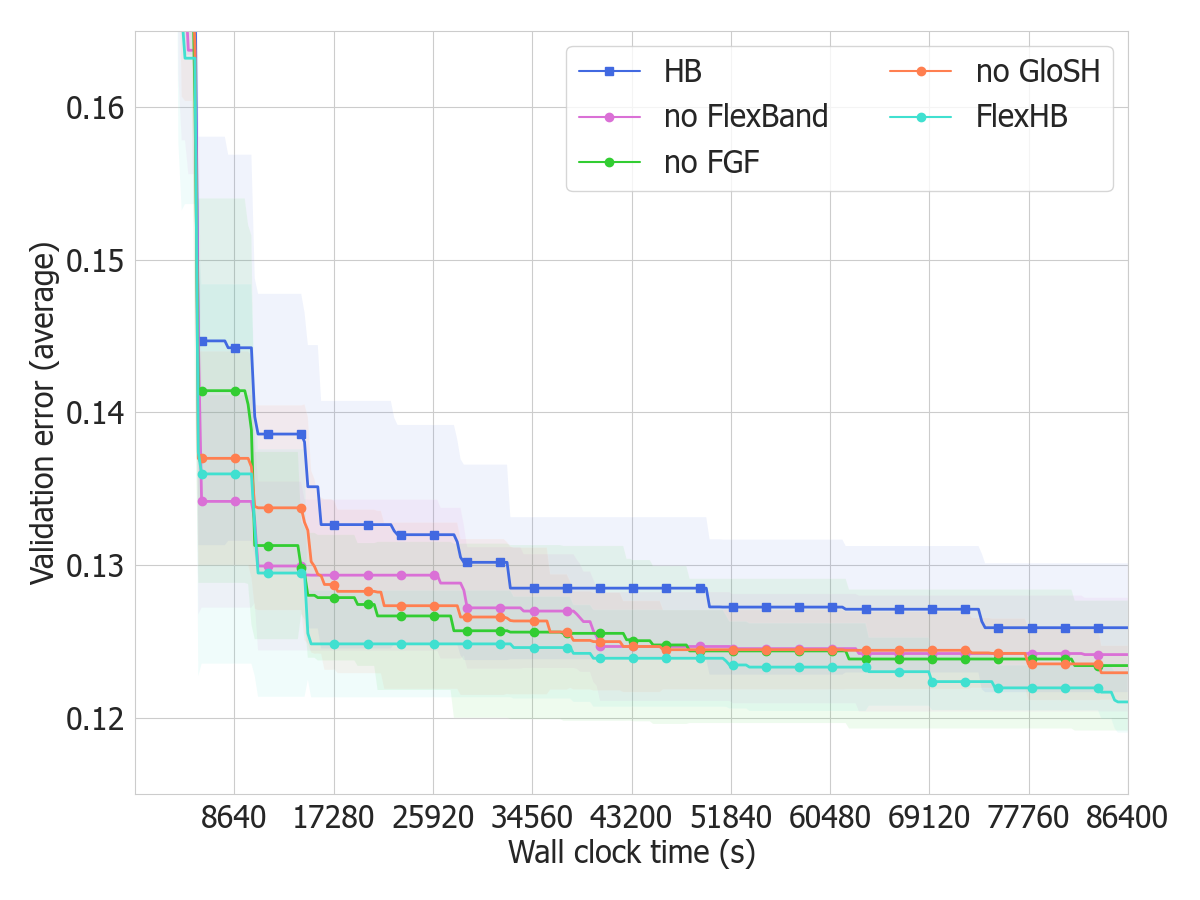}
         \caption{ResNet on CIFAR10}
         \label{fig:abl ResNet}
     \end{subfigure}
     \hfill
     \begin{subfigure}[b]{0.33\linewidth}
         \centering
         \includegraphics[width=\linewidth]{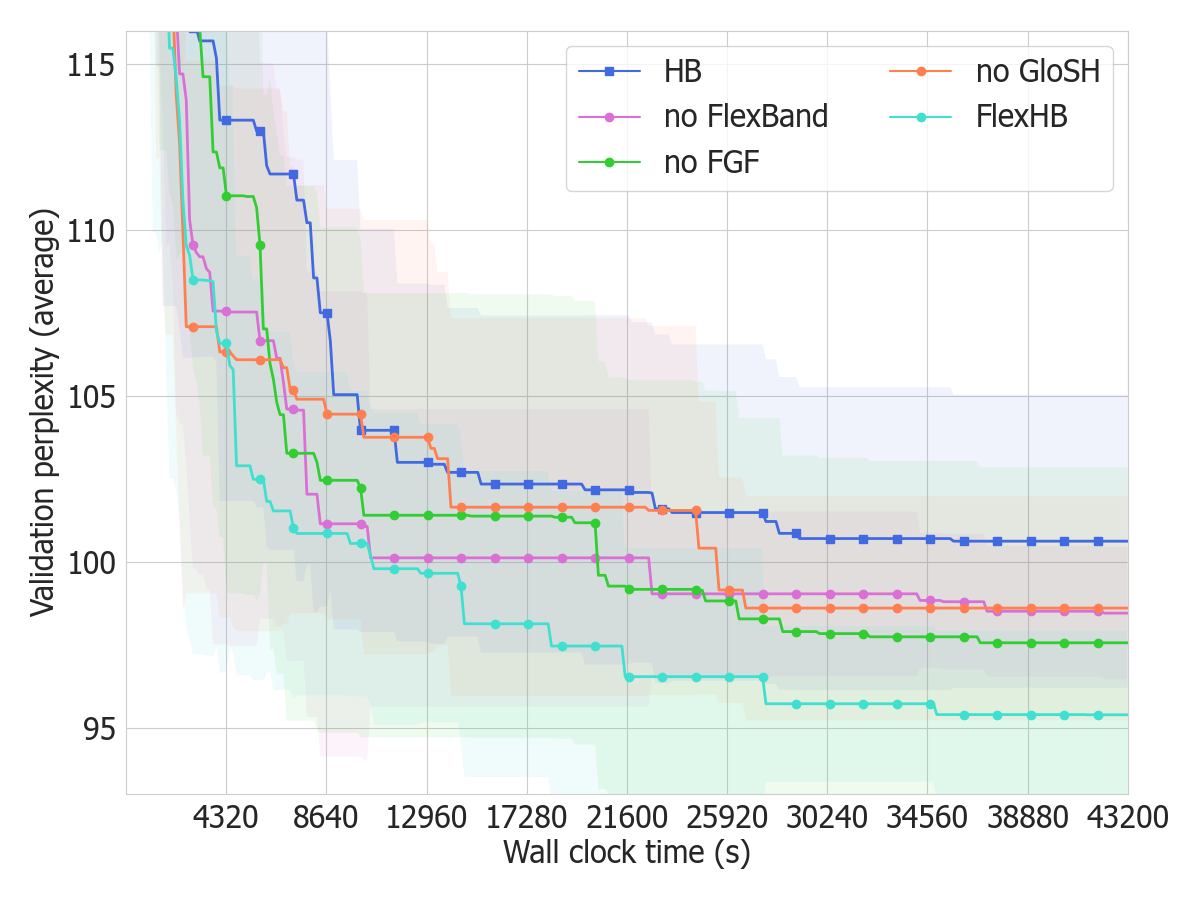}
         \caption{LSTM on Penn Treebank}
         \label{fig:abl LSTM}
     \end{subfigure}
     \caption{Results for ablation studies.}
     \label{fig:ablation visualized}
     % \vspace{-3mm}
\end{figure*}

Table \ref{tab:xgboost result full} has shown the detailed final metrics of HPO methods on XGBoost tasks, and in Figure \ref{fig:XGBoost Full Results} we display the tuning performance curves for Pokerhand, Adult and Bank dataset. 
MFES losts its advantage over BOHB on Adult and Bank tasks, and we owe its failure to the fact that these XGBoost tasks use 
subsets of whole training set as the resource unit, and the Bank and Adult dataset is not large enough. 
Therefore, lower fidelity measurements($r=1,3,9$, indicating that only 3.7\%, 11.1\% or 33.3\% of data samples are fed to model) can be skewed and lead to a low-quality ensemble of surrogates. FlexHB surmounts this drawback and obtains best final validation error.
% \vspace{-3mm}
\begin{table}[h]
% \vskip 0.1in
\centering
\resizebox{0.8\columnwidth}{!}{%
\begin{tabular}{lccccc}
\toprule
Methods    & Covtype & Pokerhand & Bank & Adult \\ \midrule
RS         &  4.17 &3.38 & 8.11 & 12.58  \\
SMAC       &  3.15 &1.64 & 8.03  & 12.40   \\
Batch-BO    &  3.12 &1.58 & 8.01  & 12.33   \\
HB         &  3.97 &3.07     & 8.09  & 12.48   \\
BOHB       &  3.35 &2.31    & 8.03  & 12.36  \\
TSE        &  3.72 & -      &  -         & -     \\
MFES       &  3.16 &1.52 & 8.04  & 12.35  \\
Our Method &  \textbf{3.09} & \textbf{1.31} &  \textbf{8.00} & \textbf{12.32}       \\
\bottomrule
\end{tabular}%
}
\caption{Validation error (\%) of all methods on XGBoost tasks, mean values on 10 times of experiments are listed. TSE is applied to Covertype dataset only. BOMS and BOCF are not applicable.}
\label{tab:xgboost result full}
% \vskip -0.1in
\end{table}
\subsection{C.3 Details of 2D Hyperparameter Space}
\label{appendix: 2d hspace}
The 2D hyperparameter space for tuning (Figure \ref{fig:Main Glosh} and \ref{fig:vanillaSHvsGloSH}) is obtained by keeping only two hyperparameters in the MLP task, including sizes of two hidden layers respectively. The value range for hyperparameters are chosen to form a 32*32 space. More specifically, size of each hidden layer varies from 16 to 528, with minimal gap being 16.

\subsection{C.4 Details of Fabricated Toy Benchmark}
\label{appendix:arrangements}
\begin{figure}[h]
    % \vspace{-3mm}
     \centering
     \begin{subfigure}[b]{0.49\columnwidth}
         \centering
         \includegraphics[width=\textwidth]{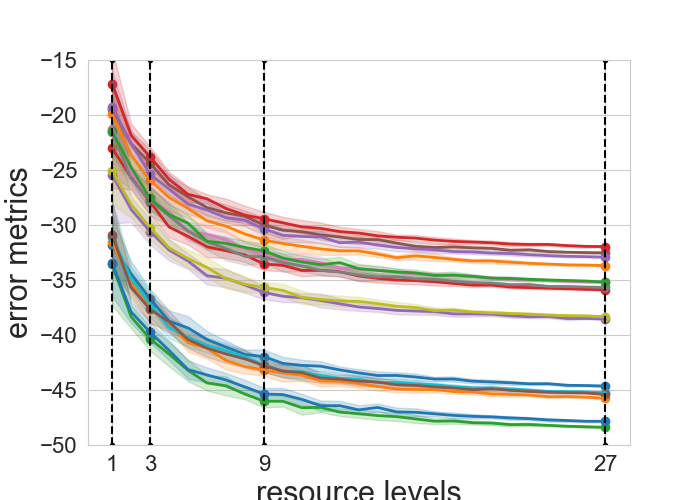}
         \caption{noise level $\phi$ = 0.5}
         \label{fig:noise_level05}
     \end{subfigure}
     \hfill
     \begin{subfigure}[b]{0.49\columnwidth}
         \centering
         \includegraphics[width=\textwidth]{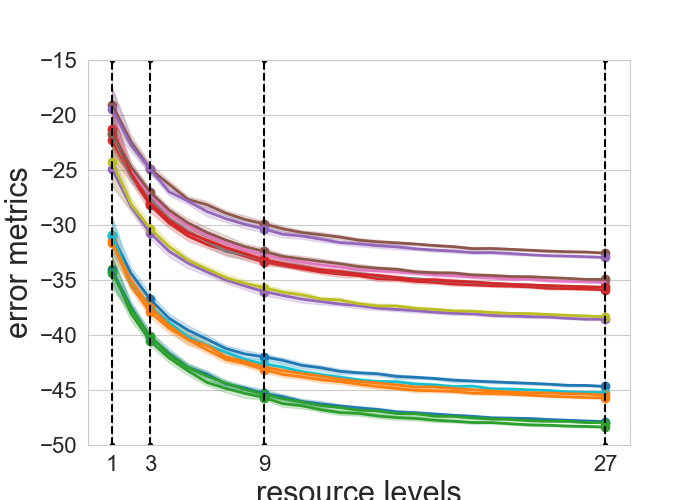}
         \caption{noise level $\phi$ = 0.2}
         \label{fig:noise_level02}
     \end{subfigure}
     % \hfill
     % \begin{subfigure}[b]{0.49\linewidth}
     %     \centering
     %     \includegraphics[width=\textwidth]{curve_noise_level05.png}
     %     \caption{curve noise level = 0.5}
     %     \label{fig:curve_noise_level05}
     % \end{subfigure}
     \caption{Visualization for learning curves with different noise levels, $x$-axis for epoch number, $y$-axis for validation error.}
     \label{fig:noise level visualzed}
     % \vspace{-3mm}
    %  \vskip-0.1in
\end{figure}

We use a toy benchmark (a manually-designed hyperparameter space) to demonstrate HPO tasks with different variance levels. The hyperparameter space is 3-dimension, including three hyperparameters to tune, namely $x, y$ and $z$. The value range of $x, y, z$ is $[-10, 10]$, $[0, 20]$ and $[0, 10]$.
The validation error $t$ to minimize during tuning is calculated by the 
following formula \eqref{eq: epoch bias} and \eqref{eq: validation error fabricated}:
\begin{equation}
    b = \frac{-20.02}{1+(\frac{epoch}{2.569})^{1.171}} + 32.935
    \label{eq: epoch bias}
    % \vspace{-1mm}
\end{equation}
\begin{equation}
    t = -scale(\frac{x^2 + (\frac{y}{4})^2 + z}{x_{max}^2 + (\frac{y_{max}}{4})^2 + z_{max}}) - b + noise(epoch, \phi)
    \label{eq: validation error fabricated}
\end{equation}
$b$ represents the learning process after epochs, 
$scale$ will re-scale the value 
in a min-max scaling way, simulating the upper and lower bound of 
the potential of all hyperparameter configurations. 
Further, $noise$ is a function that samples random noise from 
Gaussian $\mathcal{N}(0,\,\sigma^{2})$ where $\sigma$ is calculated 
based on $epoch$ and $\phi$, larger $epoch$ or smaller $\phi$ results 
in smaller $\sigma$. Given this toy benchmark, 
Figure \ref{fig:noise level visualzed} shows learning curves 
(validation error vs the number of resources) of randomly 
sampled configurations under two different noise level $\phi$, we can tell that 
larger noise level results in more turbulent learning curves. 
Performance of tuning for these two cases are shown in Figure 
\ref{fig:noise level comparison} of 
% Section \ref{subsec: Experiments of Arrangements}, 
this paper, 
we conduct experiments for 10 times to obtain results, i.e., the mean of validation metrics and the corresponding standard error.

\subsection{C.5 Details of Ablation Study}
\label{appendix: ablation}
\begin{table}[H]
    % \vskip 0.1in
    \centering
    \resizebox{0.75\linewidth}{!}{%
    \begin{tabular}{llll}
    \toprule
    Method      & MLP & ResNet & LSTM \\
    \midrule
    HB          &7.56& 12.59 &100.6      \\
    no FGF      &7.42& 12.34 &97.6 \\
    no GloSH    &7.29& 12.30 &98.6  \\
    no FlexBand &7.32& 12.41 &98.5      \\
    FlexHB      &\textbf{7.23}& \textbf{12.10} &    \textbf{95.4} \\
    \bottomrule
    \end{tabular}%
    }
    \caption{Converged validation metric of methods in ablation studies. Validation error(\%) for MLP and ResNet tasks, perplexity for LSTM. }
    \label{tab:abl full metrics}
    % \vskip -0.1in
\end{table}
As previously mentioned, methods in our ablation study including:\\
(1) no FGF: vanilla multi-fidelity BO(as in MFES-HB) used as sampling strategy, combined with FlexBand and GloSH.\\
(2) no GloSH: FGF method combined with FlexBand, use SH brackets instead of GloSH.\\
(3) no FlexBand: FGF method combined with HyperBand, use GloSH brackets instead of SH.
% (4) FlexHB: use all introduced components.

We tune MLP, ResNet and LSTM tasks with these methods above and detailed results are shown in Table \ref{tab:abl full metrics} and Figure \ref{fig:ablation visualized}. As in other experiments, we conduct tuning for 10 times and show the average value (and variance) of metrics. As Figure \ref{fig:abl MNIST} indicates, FGF contributes a lot for the performance on the MLP(MNIST) task, and FlexBand improves the anytime performance especially after the beginning stage of tuning. For the LSTM task (Figure \ref{fig:abl LSTM}), GloSH plays a critical role compared to other improvements. FlexHB, by combining FGF, GloSH and FlexBand, achieves better efficiency than any other methods on all tasks.
% \vspace{-3mm}
\subsection{C.6 Experiments on LCBench}
\label{appendix: LCBench}
LCBench \cite{LCBench} contains the learning curves 
(training or validation metrics for 50 epochs) of 2000 configurations over 35 datatsets. 
In our experiments, 
we   
utilize the validation accuracy for each epoch, HPO methods are set to seek the 
configuration with lowest validation error (1.0 minus validation accuracy). 

Our configuration 
space is as predefined in LCBench. 
For SH-like methods (HyperBand, BOHB, MFES-HB, FlexHB, etc), $\eta=3$. For FlexHB, $g$ is equal to $\eta$. 
We set the resource unit as one epoch and $R=27$. 
All experiments are conducted for 10 times with various random seeds, and 
the mean of metrics and the standard error of the mean are reported. 
Time limit for tuning 
is 750 seconds since retrieving metric of a configuration by 
LCBench's released 
API\footnote{\url{https://github.com/automl/LCBench}} 
can be done almost instantly. 
To simulate the time cost for training a real-world ML model, 
% tuning are set to wait for $0.5$ second after each epoch, 
each epoch are set to wait randomly for $0.3$ to $0.7$ second to retrieve the metric, 
so that a full fidelity candidate ($R=27$) 
needs $\geq 8.1$ seconds to return its final metric to the running HPO method. 

\begin{figure}[h]
    % \centering
    \begin{subfigure}[b]{0.5\linewidth}
        \includegraphics[width=\linewidth]{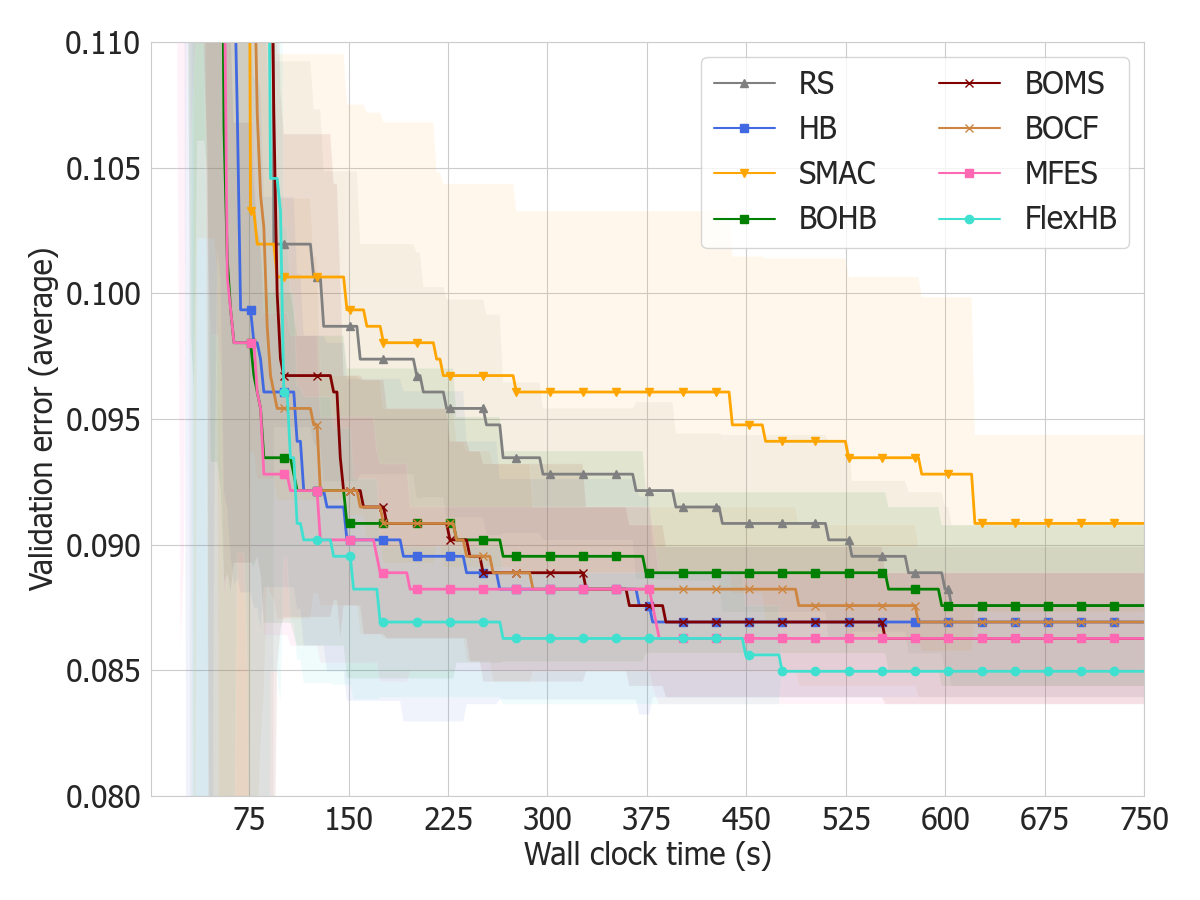}
        \caption{Australian}
        \label{LCBench1}
    \end{subfigure}%
    \begin{subfigure}[b]{0.5\linewidth}
        \includegraphics[width=\linewidth]{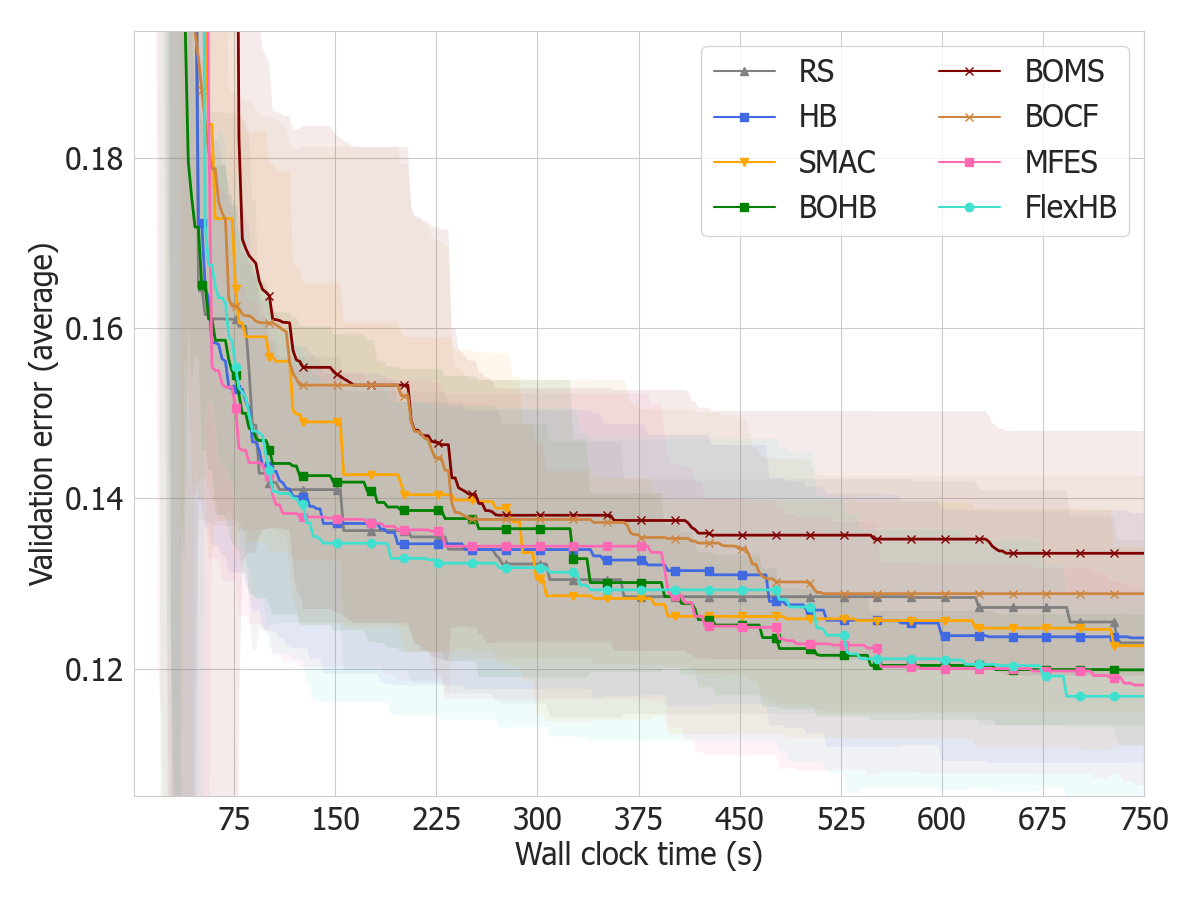}
        \caption{Fashion-MNIST}
        \label{LCBench2}
    \end{subfigure}%
    \vskip\baselineskip

    \begin{subfigure}[b]{0.5\linewidth}
        \includegraphics[width=\linewidth]{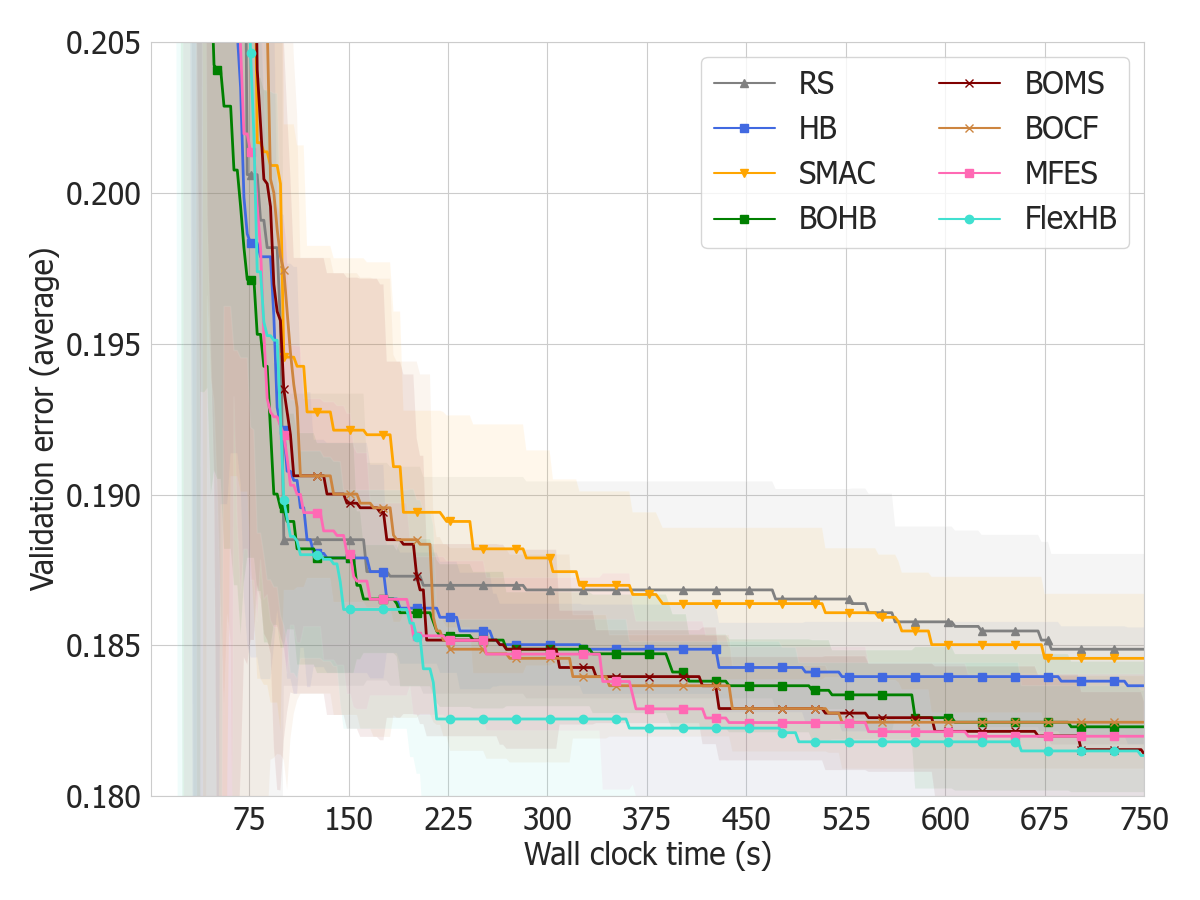}
        \caption{jasmine}
        \label{LCBench3}
    \end{subfigure}%
    \begin{subfigure}[b]{0.5\linewidth}
        \includegraphics[width=\linewidth]{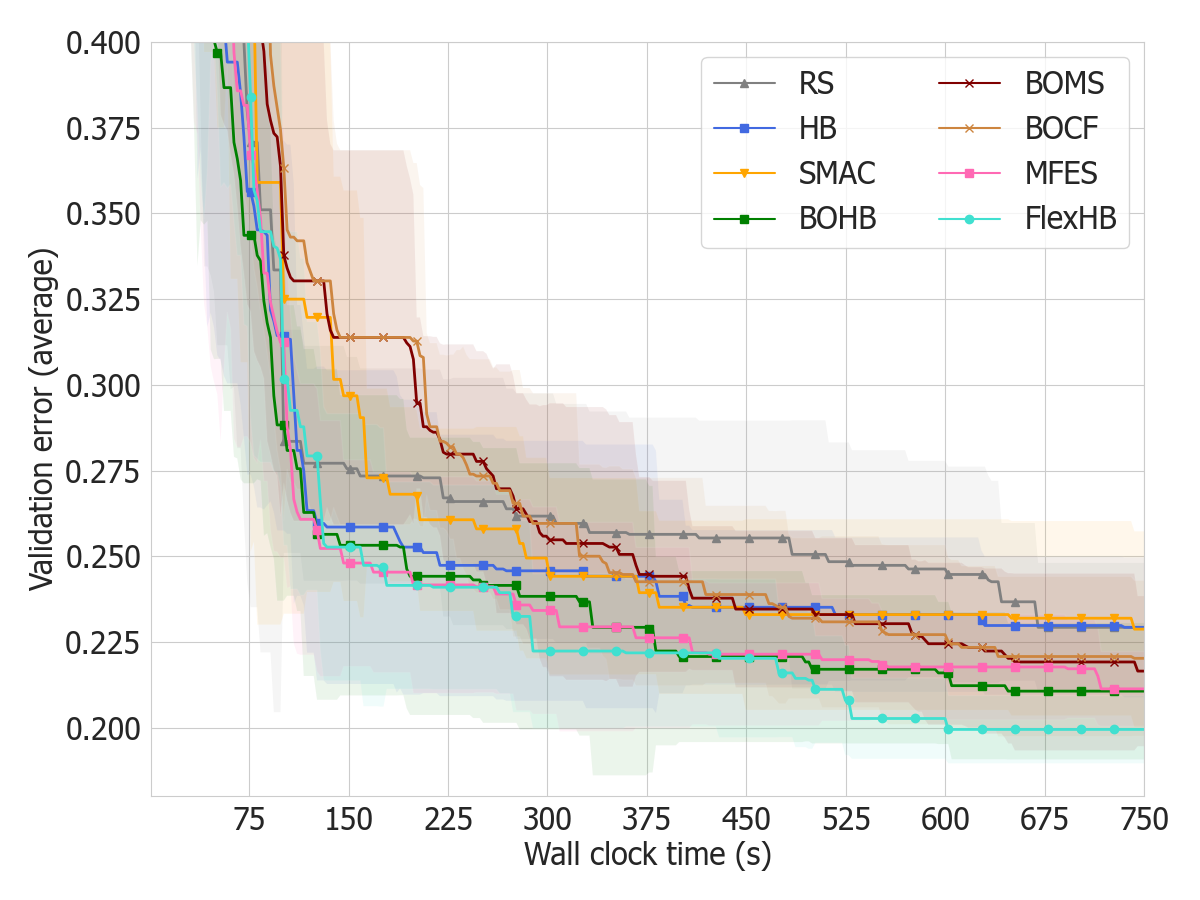}
        \caption{vehicle}
        \label{LCBench4}
    \end{subfigure}%
    \vskip\baselineskip

    \begin{subfigure}[b]{0.5\linewidth}
        \includegraphics[width=\linewidth]{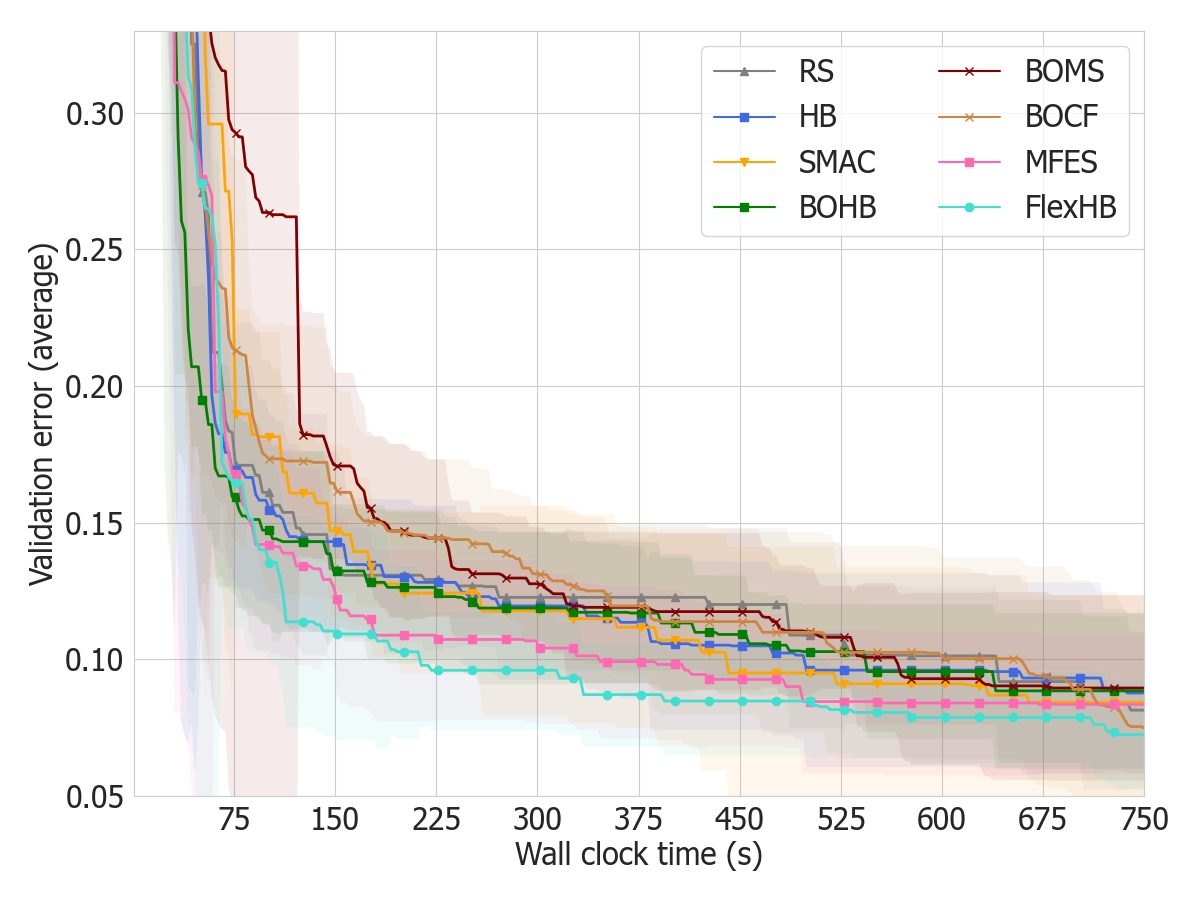}
        \caption{car}
        \label{LCBench5}
    \end{subfigure}%
    \begin{subfigure}[b]{0.5\linewidth}
        \includegraphics[width=\linewidth]{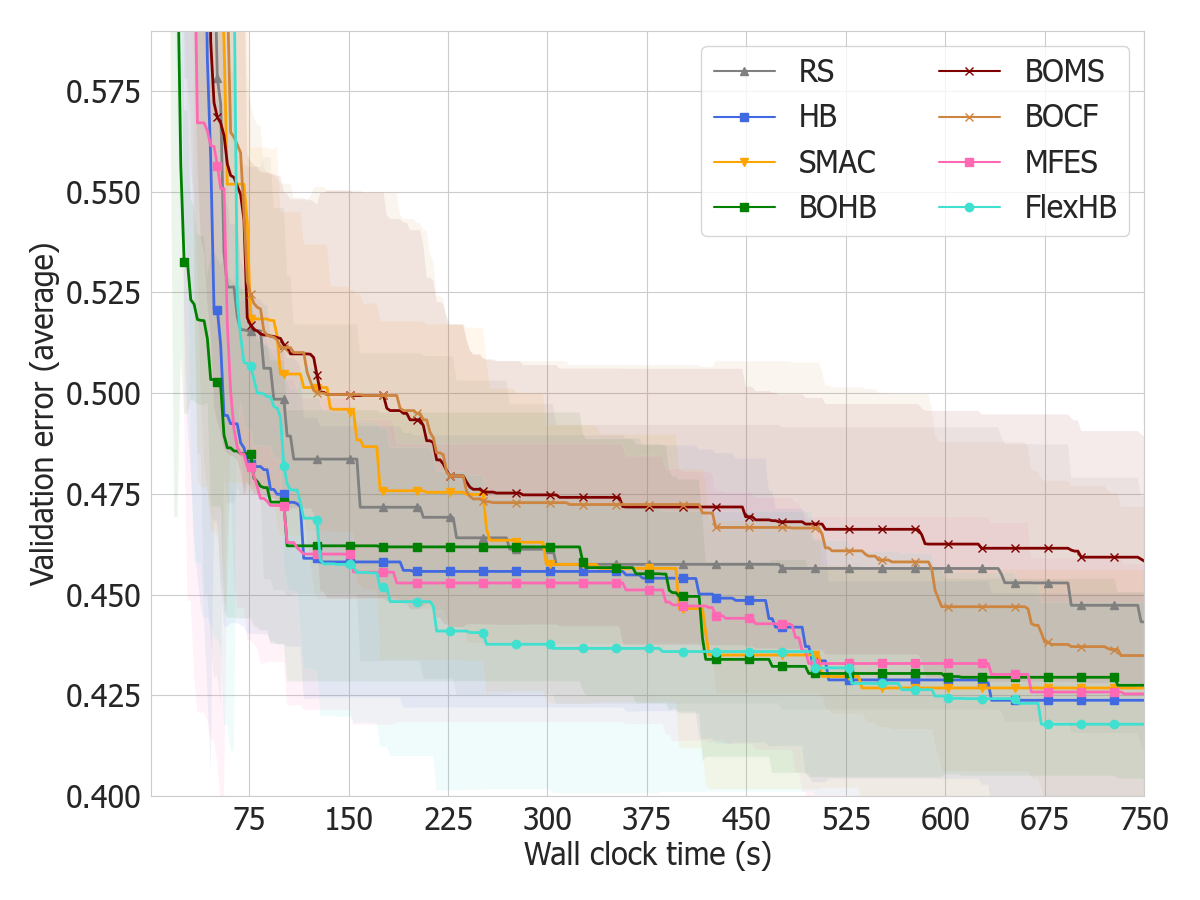}
        \caption{volkert}
        \label{LCBench6}
    \end{subfigure}%

    \caption{Results for LCBench}
    \label{fig:LCBench figure}
    % \vspace{-3mm}
\end{figure}

We mainly conduct HPO tuning on 6 datasets in LCBench and summarize the 
HPO performance in 
% Table \ref{tab:LCBench table} and 
Figure \ref{fig:LCBench figure}. 
FlexHB ranks 
as the top method, followed by MFES-HB, BOHB and HB. 
It is also notable that FlexHB is more likely to achieve (near-)optimal 
configuration area than other methods in the early stage (before 375 seconds) of tuning. 
In summary, FlexHB achieves great efficieny even compared with other popular methods. 

% \begin{table}[h]
%     % \vskip 0.1in
%     \centering
%     \resizebox{0.8\columnwidth}{!}{%
%     \begin{tabular}{lccccc}
%     \toprule
%     Methods    & Airlines & Australian & Car & Adult & FashionMNIST & Helena & Jasmine & Nomao & Vehicle & Volkert \\ \midrule
%     RS         &  X & X & X & X  \\
%     SMAC       &  X & X & X  & X   \\
%     Batch-BO    &  X & X & X  & X   \\
%     HB         &  X & X     & X  & X   \\
%     BOHB       &  X & X    & X  & X  \\
%     TSE        &  X & -      &  -         & -     \\
%     MFES       &  X & X & X  & X  \\
%     Our Method &  \textbf{X} & \textbf{X} &  \textbf{X} & \textbf{X}       \\
%     \bottomrule
%     \end{tabular}%
%     }
%     \caption{Validation error (\%) of all methods on XGBoost tasks, mean values on 10 times of experiments are listed. TSE is applied to Covertype dataset only. BOMS and BOCF are not applicable.}
%     \label{tab:LCBench table}
%     % \vskip -0.1in
% \end{table}

\end{document}